\documentclass[sigconf]{acmart}
\renewcommand\footnotetextcopyrightpermission[1]{} 
\makeatletter
\renewcommand\@formatdoi[1]{\ignorespaces}
\makeatother

\usepackage{booktabs} 
\usepackage{bbm} 
\usepackage{mathtools}
\usepackage{amsfonts}
\usepackage{graphicx}
\usepackage{xcolor}
\newcommand{\eat}[1]{}

\makeatletter
\renewcommand\section{\@startsection{section}{1}{\z@}%
	{-8\p@ \@plus -4\p@ \@minus -4\p@}%
	{6\p@ \@plus 4\p@ \@minus 4\p@}%
	{\normalfont\large\bfseries\boldmath
		\rightskip=\z@ \@plus 8em\pretolerance=10000 }}
\renewcommand\subsection{\@startsection{subsection}{2}{\z@}%
	{-8\p@ \@plus -4\p@ \@minus -4\p@}%
	{6\p@ \@plus 4\p@ \@minus 4\p@}%
	{\normalfont\normalsize\bfseries\boldmath
		\rightskip=\z@ \@plus 8em\pretolerance=10000 }}
\renewcommand\subsubsection{\@startsection{subsubsection}{3}{\z@}%
	{-4\p@ \@plus -4\p@ \@minus -4\p@}%
	{-1.5em \@plus -0.22em \@minus -0.1em}%
	{\normalfont\normalsize\bfseries\boldmath}}
\makeatother

\AtBeginDocument{%
  \providecommand\BibTeX{{%
    \normalfont B\kern-0.5em{\scshape i\kern-0.25em b}\kern-0.8em\TeX}}}

\acmDOI{}

\begin{document}

\title{ORD: Object Relationship Discovery\\ for Visual Dialogue Generation}

\author{Ziwei Wang}
 \affiliation{%
 	\department{School of Information Technology and Electrical Engineering}
 	\institution{The University of Queensland}
 }
\email{ziwei.wang@uq.edu.au}

\author{Zi Huang}
 \affiliation{%
 	\department{School of Information Technology and Electrical Engineering}
 	\institution{The University of Queensland}
 }
 \email{huang@itee.uq.edu.au}

\author{Yadan Luo}
 \affiliation{%
 	\department{School of Information Technology and Electrical Engineering}
 	\institution{The University of Queensland}
 }
 \email{lyadanluol@gmail.com}

\author{Huimin Lu}
 \affiliation{%
 	\department{Department of Electrical Engineering and Electronics}
 	\institution{Kyushu Institute of Technology}
 }
 \email{riku@cntl.kyutech.ac.jp}


\begin{abstract}
	With the rapid advancement of image captioning and visual question answering at single-round level, the question of how to generate multi-round dialogue about visual content has not yet been well explored.
	Existing visual dialogue methods encode the image into a fixed feature vector directly, concatenated with the question and history embeddings to predict the response.
	Some recent methods tackle the co-reference resolution problem using co-attention mechanism to cross-refer relevant elements from the image, history, and the target question.
	However, it remains challenging to reason visual relationships, since the fine-grained object-level information is omitted before co-attentive reasoning.
	In this paper, we propose an object relationship discovery (ORD) framework to preserve the object interactions for visual dialogue generation. Specifically, a hierarchical graph convolutional network (HierGCN) is proposed to retain the object nodes and neighbour relationships locally, and then refines the object-object connections globally to obtain the final graph embeddings.
	A graph attention is further incorporated to dynamically attend to this graph-structured representation at the response reasoning stage.
	Extensive experiments have proved that the proposed method can significantly improve the quality of dialogue by utilising the contextual information of visual relationships. The model achieves superior performance over the state-of-the-art methods on the Visual Dialog dataset, increasing MRR from 0.6222 to 0.6447, and recall$@$1 from 48.48\% to 51.22\%.
\end{abstract}
%
%



\maketitle
\thispagestyle{empty}

\section{Introduction} \label{sec:introduction}
%
%
%
%


Multimedia content understanding has attracted increasing attention in the multimedia, vision and language fields in recent years. 
Emerging research directions such as image captioning~\cite{cap_Karpathy,mm_imgcap_binyi} and visual question answering (VQA)~\cite{vqa_iccv15_AntolALMBZP15} demonstrate promising reasoning ability towards the ultimate goal of multimedia intelligent systems.
In brief, the image captioning models describe an image with a single sentence, whilst the VQA models answer an individual question conditioned on the visual content.
However, in reality, a human is able to initiate several rounds of highly interrelated conversation about the visual content rather than a one-off response. Motivated by this intention, Visual Dialog~\cite{vd_cvpr17_das} is proposed recently as a challenging extension of captioning and VQA. In visual dialogue, a dialogue agent is built to constantly answer multi-rounds of highly coherent questions grounded in visual content and conditioned on dialogue history.
\begin{figure}[t]
	\centering
	\includegraphics[width=0.50\textwidth, trim={0.5cm 0.5cm 0.5cm 0.2cm}]{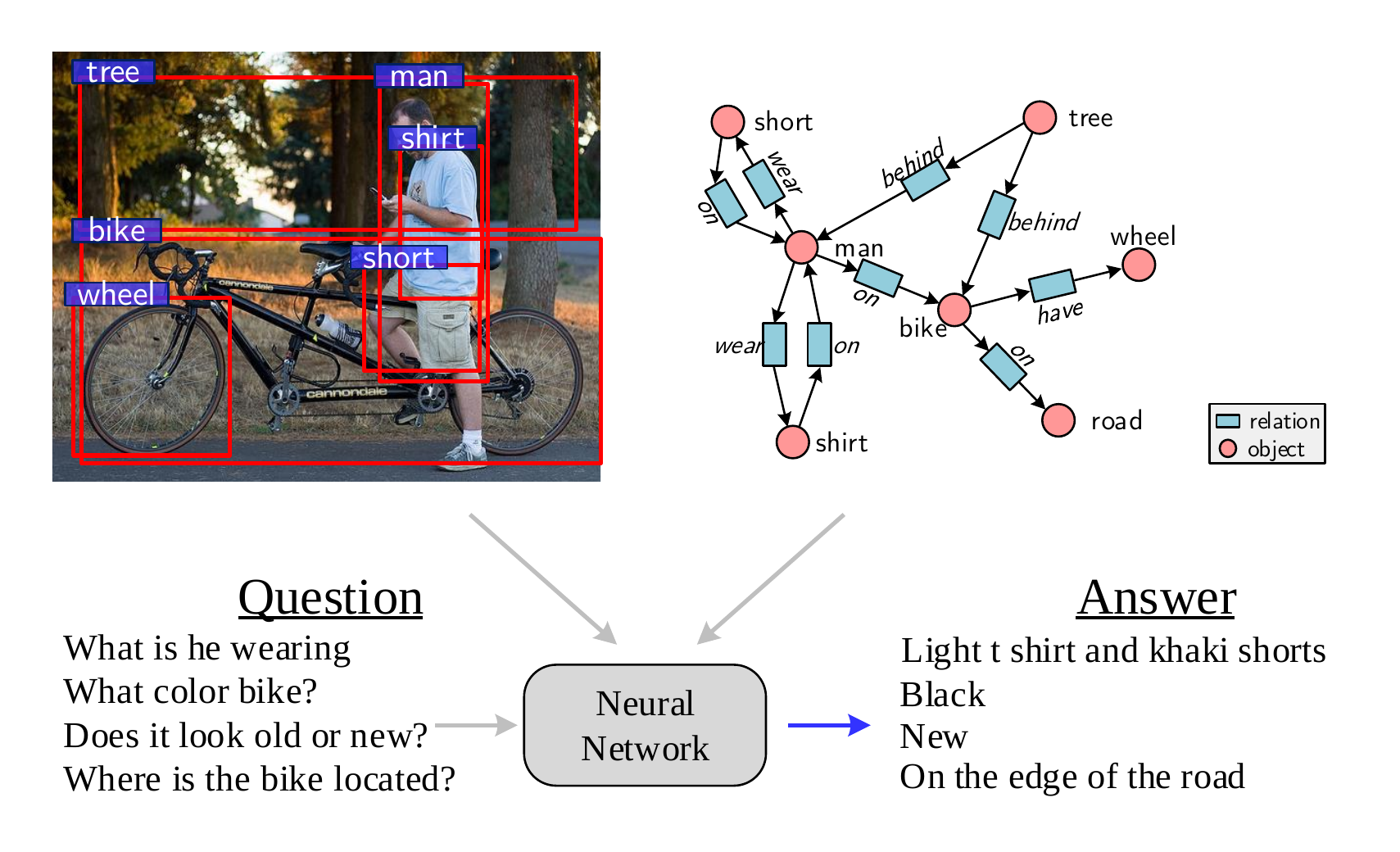}
	\caption{
		The proposed model detects the salient regions, reasons the object interactions, and encodes the visual relationships as a scene graph.
		The neural network is trained to reason over the representations of image, scene graph, question and dialogue history, and to predict a suitable answer.
	}
	\label{fig:overview_showcase}
\end{figure}

The multi-modal nature of the input question, image and history requires the dialogue model to simultaneously understand textual questions, extract salient visual features and retrieve relevant dialogue history for response prediction.
In general, the existing visual dialogue methods consist of a multi-modal encoder and a language response decoder. The basic multi-modal encoder contains three sub-modules: convolutional neural network (CNN) image encoder, question long-short term memory (LSTM) language model, and history LSTM language model. 
Some recent work tackles the co-reference resolution problems, which is a task in natural language processing to find all expressions referring to the same entity given text content. These methods utilise co-attention mechanism to cross-refer both visual and textual elements from multi-modal sources for comprehensive feature representation.

\eat{Meanwhile, following the previous work~\cite{vd_cvpr17_das}, the response decoder has two different types: discriminative and generative. 
	Aiming at reasoning clues from image and history given the question, previous work incorporates memory network~\cite{vd_cvpr17_das}, co-reference attention mechanism~\cite{vd_nips_LuKYPB17,vd_cvpr_Wu0S0H18} to enhance the recognition of major visual and textual content. Furthermore, advance learning strategies such as generative adversarial training~\cite{vd_cvpr_Wu0S0H18} and knowledge transfer learning~\cite{vd_nips_LuKYPB17} are proposed to enhance the generalisation capability of the learned model. Concurrently, another branch of dialogue systems~\cite{vd_rl_cvpr_MassicetiSDT18, vd_rl_iccv_DasKMLB17} learn the policies of both question-agent and answer-agent, and they formalise a conversation game between these two agents as a reinforcement learning (RL) problem. However, these RL-based dialogue systems are not in the scope of this paper, since the main focus of this work is reasoning the answer based on the provided question.}

Despite the fact that co-reference resolution methods have made multi-modal cues logically coupled at a high level, the fine-grained visual cues are not thoughtfully exploited. The recent methods simply use CNN visual features, without considering object regions and visual relationships, therefore lacking recognition on locative relationships between regions of interest.
The information loss in object-object interactions can be forgivable in simple scenes such as ``\textit{a pencil on the table}'', ``\textit{sheep on the grass}'', in which the number of objects is limited.
\eat{and the background noise is manageable}
However, such ``pure'' scenes are not commonly obtainable in real world scenarios. Intuitively, the real images unavoidably contain multiple instances, overlapping spatial arrangements and noisy background. For example, in Figure~\ref{fig:overview_showcase}, the picture contains many closely arranged or overlapped objects (e.g. \textit{bike-wheel}, \textit{human-shirt-short}), and a complex background with criss-cross \textit{trees}, \textit{grass} and bright \textit{sky}. 

\eat{it is still challenging to answer questions with referent like ``Does \textit{it} look old or new?'' for a basic dialogue model. Thankfully, the co-reference resolution methods are able to determine the referent of ``it''.}

Although it is trivial for a neural dialogue agent to answer na\"ive questions such as ``Is it a sunny day?'' given general visual context,
it is still challenging to answer the fine-grained object-level details such as ``What is he wearing?'', or ``Where is the bike located?''. The global visual feature and high-level co-attention methods are not sufficient to reason object attributes and visual relationships.

To alleviate the above issues, we propose an attentive graph-structured hierarchical model for visual dialogue. To the best of our knowledge, this is the first paper to leverage scene graphs in a visual dialogue task. 
The scene graph is a graph structure to represent the scene, where the nodes are objects (e.g. \textit{human},\textit{shirt}), and the edges are relationships between objects (e.g. $\langle \mathit{human-wear-shirt} \rangle$ , $\langle \mathit{shirt-on-human}\rangle$ ).
In this work, the scene graph only contains the objects and relationships without attributes following \cite{sg_LuKBL16,sg_2017_li_msdn,sg_XuZCF17, sg_fnet_2018}.
	The object attributes can be further encoded in the object embedding as auxiliary information, but we do not consider this information for simplicity.

\begin{figure*}[t]
	\includegraphics[width=1\textwidth, , trim={0.7cm 0.6cm 0.7cm 0.6cm}, clip=true]{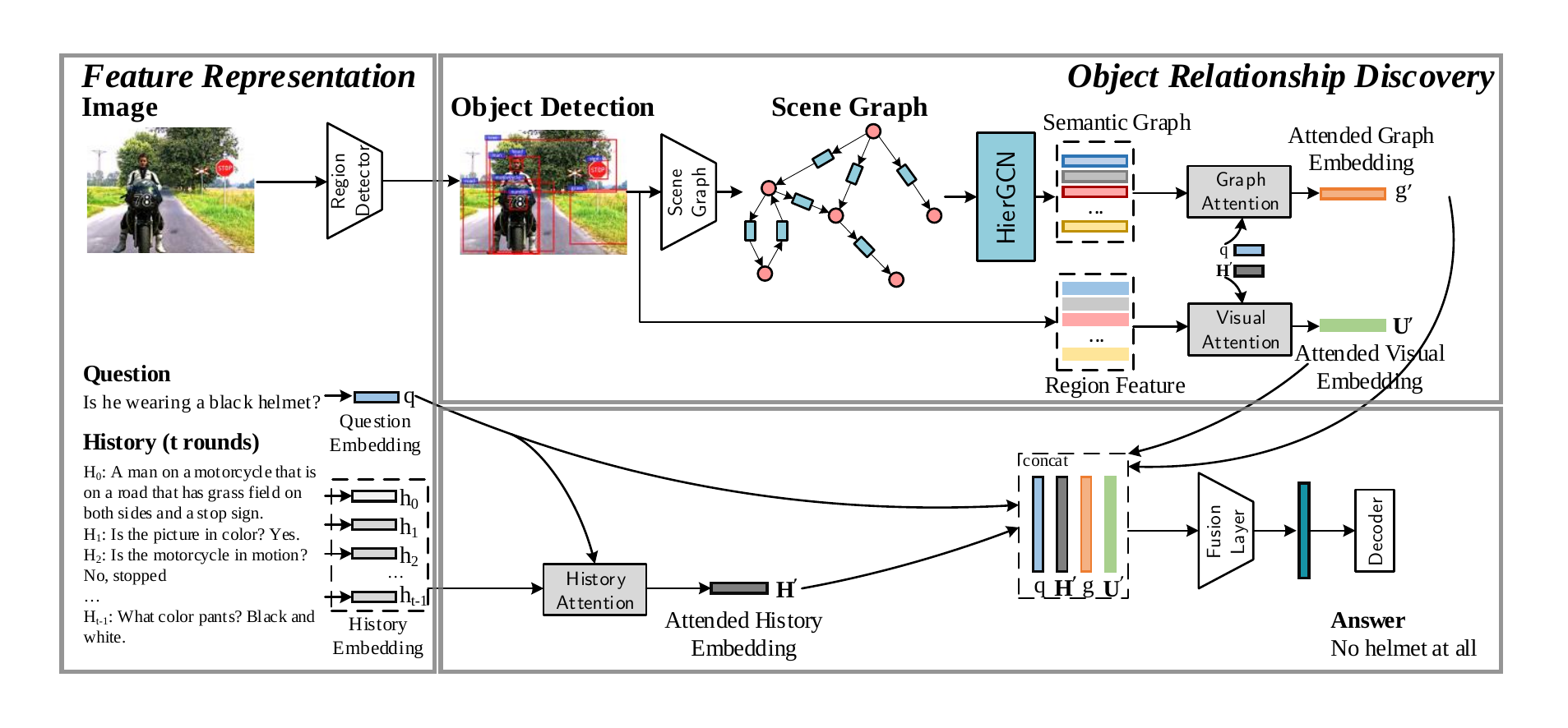}
	\centering
	\caption{Architecture of the visual dialogue generation framework.
		In general, given an image, question and the dialogue history, the proposed model encodes all the features into a combined feature representation to predict the answer.
		For image, the object regions are firstly detected from the image, followed by two embedding streams: scene graph and visual region features.
		In scene graph embedding stream, the visual relationships are detected from region proposals, and the graph structure is preserved via a novel Hierarchical Graph Convolutional Network (HierGCN). 
		A novel graph attention mechanism is designed to selectively focus on relevant object nodes conditioned on the question and attended history to manifest the attended graph embeddings.
		Similarly, the attended history and visual embeddings are obtained from the individual attention modules.
		Finally, the question and attended context embeddings are fused and forwarded for answer decoding.
	}
	\label{fig:overview_system}
\end{figure*}

The proposed model utilises the scene graph to preserve the object interactions in the image, whilst retaining the reasoning capability of co-reference resolution models. 
Figure \ref{fig:overview_system} illustrates an overview of the proposed visual dialogue framework.
Technically, the tri-stream neural network is derived as the encoder, to jointly learns the representation for image, question and history features.
In the visual stream, the proposed framework adopts Faster-RCNN~\cite{faster_rcnn} to extract regional visual features, which are utilised in both scene graph generation and visual feature representation.

To preserve visual relationships in the generated scene graph, we propose a Hierarchical Graph Convolutional Network (HierGCN) to integrate graph structure into neural visual dialogue framework. The HierGCN captures object nodes and their neighbour relationship edges locally, and then a global GCN refine the graph representations to preserve connections between objects.

\eat{Moreover, the scene graph is generated based on regional visual features, and the graph structure is preserved by the proposed Hierarchical Graph Convolutional Network (HierGCN). 
	In HierGCN, the objects and neighbour relationships are preserved using local GCN, and the entire scene is further encoded via global GCN (See details in Section~\ref{sec:method-hierGCN}).}

Furthermore, the graph attention mechanism is incorporated to capture the salient object nodes in scene graph based on the current context. 
Similarly, the history and visual attention modules are equipped to reason the relevance to the target question.
Finally, the fused representation of attended graph, visual, history features and question embedding is decoded by standard discriminative or generative decoder to predict the answer.
Different from all existing visual dialogue methods, the proposed ORD model detects fine-grained object-level regions, encodes the scene with hierarchical understanding of visual relationships, rather than simply summarise visual features from convolutional layers, and therefore significantly improves dialogue response reasoning.

The contributions in this paper are three-fold:
\begin{enumerate}
	\item To the best of our knowledge, it is a novel attempt to leverage object relationships encoded in graph representation for visual dialogue tasks, with which the generated dialogues reveal visual details, sensitivity to object-object relationships and robust to background noise.
	\item Different from previous work, the fine-grained object detection is utilised to extract details from the visual content. A novel Hierarchical Graph Convolutional Network (HierGCN) is proposed to better preserve graph-structured object interactions, and a graph attention mechanism is equipped to enable an intelligent shift among focused objects.
	\item Extensive experiments are conducted on the Visual Dialog dataset illustrating the effectiveness of the proposed scene graph-structured framework, the HierGCN and the graph attention module.
\end{enumerate}

The rest of this paper is organised as follows: Section~\ref{sec:related_work} reviews visual dialogue and object relationships detection methods. Section~\ref{sec:method} introduces a basic scene graph-structured dialogue generation model, and the full model with graph attention. Section~\ref{sec:experiment} contains experiments, and Section~\ref{sec:conclusion} concludes the proposed framework.

\section{Related Work} \label{sec:related_work}

\subsection{Visual Dialogue} 
Multimedia content understanding tasks such as captioning~\cite{cap_Karpathy,cap_VinyalsTBE15,cap_xu2015show,cap_krause2016paragraphs,cap_depth_WangLLHY18}, visual question answering (VQA)~\cite{vqa_updown,vqa_SAN_YangHGDS16,vqa_HieCoAtt_LuYBP16, aaai_vqa_gao} have been intensively studied in the multimedia community recently.
Even though understanding a visual content with a single sentence (image captioning) or a question-answer pair (VQA) has made great success, massive underlying information such as object-object interaction, geometric structure conveyed in the scene would be omitted if only condensing the visual scene into a single round of summary. 
As an extension of image captioning and VQA, multi-round visual dialogue has been proposed in~\cite{vd_cvpr17_das} to tackle the above deficiencies of a single round brief. Specifically, Das et al.~\cite{vd_cvpr17_das} introduced a visual-based dialogue task named VisDial. In VisDial, the questioner (e.g. human) asks the questions about visual content, and the dialogue agent will compose a human-understandable response based on the question, visual content and dialogue history. The visual content and the history of conversation are only held by the dialogue agent.

The emerging work mainly focuses on the dependencies between the question, image and history to solve the visual dialogue task following an encoder-decoder framework. Lu et al.~\cite{vd_nips_LuKYPB17} proposed a history-conditioned image attentive encoder (HCIAE) to attend relevant history and salient image region. This work further improves the generative decoder by transferring knowledge from the discriminative model. Following the same motivation, a sequential co-attention mechanism is proposed in~\cite{vd_cvpr_Wu0S0H18}. The co-attention jointly generates attention maps between the input question, image and history, and all the co-attended features are combined into the final feature as the encoder output in the final stage. An adversarial learning strategy is adopted to self-assess the quality of response while learning, which further improves the performance. A recent graph neural network (GNN) based method~\cite{vd_Zheng_2019_CVPR} reasons the dependencies between textual question and answers. However, different from this GNN method, the focus of the proposed work is on the visual relationships underlying in the image.

In general, existing visual dialogue methods~\cite{vd_kottur2018visual, vd_niu2019recursive} highly rely on the high-level co-reference resolution, knowledge transfer and adversarial training strategy. 
The underexploited visual details (e.g. region features, object-object interactions) inevitably make generated answers vague and lack discrimination on relative spatial relationships between objects. The proposed ORD, on the other hand, is to enrich the feature representation to generate more meaningful embedding by modelling the scene as a structured graph with relationships to facilitate the logical reasoning process effectively.

\subsection{Visual Relationships Detection} 

The visual relationships between objects have received increasing attention recently~\cite{sg_XuZCF17, sg_2017_li_msdn,mm_rel_HTShen}, which have proven to be beneficial to varieties of vision tasks~\cite{ sg_JohnsonKSLSBL15,vqa_graph_TeneyLH17,cap_sg_YaoPLM18,sg_JohnsonGF18}.

In general, the visual relationship detection methods could be divided into two trends: category-specific and generic. Early category-specific relation detectors exclusively target a specific category of relations, such as spatial relations~\cite{sg_3d_ChoiCPS13,sg_GuptaD08} and actions~\cite{sg_action_GuptaKD09,sg_action_GkioxariGM15,sg_action_YaoF10}. These models utilise visual features and geometry, but could hardly achieve satisfactory performance due to the fact that the nature of relationships does not limit to a single category, whilst the generic visual relationship detection~\cite{sg_LuKBL16, sg_DaiZL17, sg_XuZCF17} intends to predict various types of relationships with promising performance. Lu et al.~\cite{sg_LuKBL16} firstly introduced a generic visual relationship task, while this paper considers object detections as the first stage, then recognise predicates between objects with language prior. Furthermore, Xu et al.~\cite{sg_XuZCF17} formalised the visual predicates into a scene graph proposed in~\cite{sg_JohnsonKSLSBL15}. This model explicitly generates a scene graph representation from an image by an iterative message passing algorithm.
Since the proposed visual dialogue model is agnostic to scene graph generation methods, this paper adopts the latest state-of-the-art subgraph-based scene graph generation framework named Factorizable Net~\cite{sg_fnet_2018} to predict scene graph for visual dialogue.

\section{The Proposed Approach} \label{sec:method}
%
%

In this section, we introduce the proposed Object Relationship Discovery (ORD) framework to generate visual dialogue by explicitly leveraging scene graph to preserve visual relationships. As shown in Figure~\ref{fig:overview_system}, ORD firstly detects object regions (e.g. Faster-RCNN~\cite{faster_rcnn}) from the input RGB image, the detected region features are further forwarded into scene graph embedding and visual embedding branches.
In graph embedding channel, the scene graph is constructed based on the detected object regions to preserve semantic relationships (e.g. $ \langle \mathit{man-wear-shirt} \rangle$, $ \langle \mathit{tree-behind-bike} \rangle$) before encoding by the novel Hierarchical Graph Convolutional Networks. In HierGCN, the local GCN firstly captures local relationships between the object nodes and neighbour relationship edges, then the global GCN refines the node features to preserve object-object connections. Moreover, the graph attention dynamically focus on the most relevant nodes given question and attended history to generate the attended graph embedding.
Simultaneously, to retain regional visual content, the visual attention is engaged to attend salient regions based on context.
Finally, the attended graph, visual and history embeddings, combined with the question embedding, are concatenated and forwarded to fusion layer. The fused feature is the output of encoder, which is decoded by discriminative or generative decoder to obtain the dialogue response.

\subsection{Problem Formulation} \label{sec:method-formulation}

	In this section, we formally define the visual dialogue generation task as introduced by Das et al.\cite{vd_cvpr17_das}, followed by a brief introduction of the conventional framework.

	Formally, we denote the input RGB image as $ I $,
	the ground-truth multi-round dialogue history till round $t-1$ (including an image caption $ c $) as
$ \mathbf{H}  =  ( \underbrace{\vphantom{(c)}c}_{\mathit{H_0}}, \underbrace{(q_1, a_1)}_{\mathit{H_1}},  \cdots, \allowbreak \underbrace{(q_{t-1}, a_{t-1})}_{\mathit{H_{t-1}}}) $,
where the $ \{q_1, q_2, \cdots,\allowbreak q_{t-1}\} $ and $ \{a_1, a_2, \cdots,\allowbreak a_{t-1}\} $ are questions and answers until the round $t-1$ in the history, respectively.

	And follow-up target question at round $t$ is denoted as $ q_t $. The objective of the proposed model is to return a valid answer $ a_t $ given $ I $, $ \mathbf{H} $ and $ q_t $:
	\begin{equation}
		\theta^*=\arg \underset{\theta}{max}\log p(a_t|I;\mathbf{H};q_t),\label{eq:obj_basic}
	\end{equation}
	where $\theta$ is the model parameters. The encoder-decoder model is expected to return most possible answer given the question, image and dialogue history.

Following~\cite{vd_cvpr17_das}, the visual dialogue model is trained to return a sorting of 100 candidate answers 
$ \allowbreak \mathcal{A}_t = \{a_t^{(1)}, a_t^{(2)}, \cdots, \allowbreak a_t^{(100)}\}$.
Given the problem setup, two different settings are introduced: discriminative and generative. 
For discriminative setting, 
$ I $, $ \mathbf{H} $ and $ q_t $ are encoded into a combined feature representation .
	Based on $E_t$, the discriminative decoder directly gives a ranked list of 100 candidate answers, in which the top ranked candidate is chosen as the response.
The metric-learning multi-class N-pair loss~\cite{vd_nips_LuKYPB17} is adopted to train the model to maximise ground truth answer score, whilst encouraging the model to score options that are similar to the ground truth higher than the dissimilar ones.
For generative setting, the combined feature $E_t$ is extracted using same encoder as in the discriminative setting, but the generative decoder is a word sequence generator which can generate open-ended answer. For evaluation purpose, the recurrent decoder uses log likelihood scores of all the candidate answers and ranks the candidate answers based on the scores.

	In both discriminative and generative settings, the quality of the combined feature representation $E_t$ is critical to the dialogue generation task. 
	In the basic Late Fusion~\cite{vd_cvpr17_das} framework, the global image CNN features, the question representation from last hidden state of question LSTM, and the history representation from last hidden state of history LSTM are simply concatenated and linearly transformed into a combined feature $E_t$. In this fashion, the semantic relationships between visual and textual can hardly be captured. The following work~\cite{vd_nips_LuKYPB17} propose a history-conditioned image attentive encoder (HCIAE) model to perform co-reference resolution using co-attention mechanism.
	The co-attention mechanism is able to re-fine history and visual features by only focusing on the relevant context. Theoretically, the re-fined features will only contain most relevant visual and textual information for the given question. However, in practice, the image features from convolutional layers are not sufficient to eliminate background noise, and the object regions and relationships cannot be explicitly represented.

\subsection{Object Relationship Discovery} \label{sec:method-ord}
	The recent methods only consider CNN visual features extracted from the full image, without exploiting the fine-grained object regions and relationships. 
	
	In this section, we propose an Object Relationship Discovery (ORD) framework to exploit scene graph $ \mathcal{G} $ for visual reasoning.
	Specifically, we explicitly recognise object regions from the full image, and identify the relationships between objects to construct a scene graph. 
	Since the graph structure is difficult to be directly integrated in the existing visual dialogue framework, we propose a novel Hierarchical Graph Convolutional Network (HierGCN) to preserve the graph structure. And the relationship-aware graph embeddings are further attended via a novel graph attention module. 
	
	To exploit the scene graph $\mathcal{G}$ in the proposed framework, the objective function Equation~\ref{eq:obj_basic} could be rewritten as:
	\begin{equation}
		\theta^*=\arg \underset{\theta}{max}\log p(a_t|I;\mathcal{G};\mathbf{H};q_t).\label{eq:obj_graph}
	\end{equation}
	Correspondingly, the combined representation $E_t$ will preserve image, history, question, and scene graph features. The following sub-sections will introduce the scene graph generation, the Hierarchical GCN, the graph attention module, and the co-attention encoder in details.

\subsubsection{Preliminary Work on Scene Graph Generation} \label{sec:method-preliminary}
In this section, we briefly introduce how to extract image region features $ \mathbf{U} $ and build scene graph $ \mathcal{G} $ based on the given image $ I $.
First, Faster-RCNN is utilised to detect object regions in the image $I$. The region features 
$ \mathbf{U} = [u_1, u_2, \cdots, \allowbreak u_k ] $ with top-$k$ salient regions of objects in the image $I$ are extracted, where 
$ u_i \in \mathbb{R}^{D_u} $ denotes the $D_u$-dimensional visual feature of each object region.
Next, the scene graph $ \mathcal{G} = (\mathcal{V},  \mathcal{E})$ is constructed based a set of object nodes (vertices) $ \mathcal{V} $ and relationship edges $ \mathcal{E} $. 
As mentioned in Section~\ref{sec:introduction}, in this work, we do not consider object attributes in the scene graph.

To build the scene graph, the scene graph generation model (e.g. F-Net~\cite{sg_fnet_2018}) groups the detected regions in pairs, and constructs the fully-connected graph, in which each pair of object nodes is connected with directed edges (\textit{in} and \textit{out}). This model further represents the numerous relationships with fewer sub-graphs and object features to reduce computational cost, and the message are passed through the graph to refine the feature for more accurate relationships predictions. Finally, the top-$m$ $\langle \mathit{subject-predicate-object} \rangle$ relationships are kept for final scene graph construction.

\subsubsection{Hierarchical Graph Convolutional Networks for Scene Graph Embedding} \label{sec:method-hierGCN}
In this section, we discuss how to learn the graph embedding to integrate the scene graph in the proposed ORD framework.

\begin{figure}[!htb]
	\includegraphics[width=0.5\textwidth, trim={0.5cm 0.8cm 0.5cm 0.8cm}]{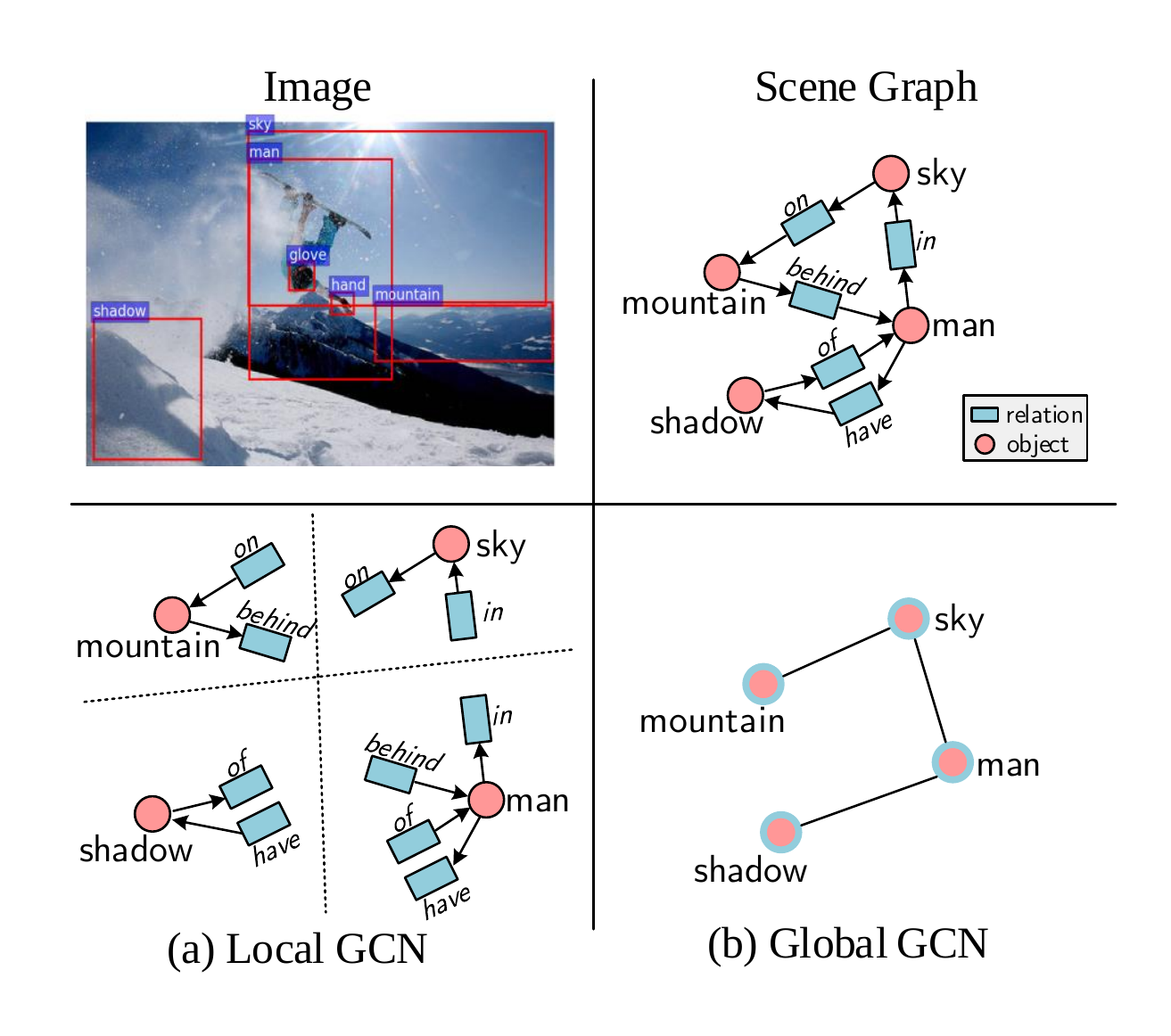}
	\centering
	\caption{An illustration of the Hierarchical Graph Convolutional Network. Given the above image and its scene graph, the Hierarchical GCN preserves the directed graph with edge features in two stages: (a) Local GCN. The nodes and their neighbour relationship nodes are fed into local GCN to preserve local graph structure. (b) Global GCN. The relationship-aware nodes are embedded via global GCN to maintain object-object connection. The local and global GCN are trained end-to-end to preserve both local neighbour structure and global scene-level node connections.}
	\label{fig:heir_gcn}
\end{figure}

Inspired by Graph Convolutional Network~\cite{gcn_KipfW17} for node classification, we propose a Hierarchical Graph Convolutional Networks (see Figure~\ref{fig:heir_gcn}) to refine the region features by preserving relationships between objects.  
We argue that the original GCN~\cite{gcn_KipfW17} can only directly use in the case of \textit{undirected} graph \textit{without edge} features, which is not sufficient in our \textit{directed} scene graph with \textit{relationship} edges. To overcome the above issues, the proposed hierarchical GCN captures the connections between the object nodes and the neighbour relationship nodes locally, and then refines the relationship-aware node connections in the scene-level globally.

	\textbf{(a) Local GCN}. 
	The local GCN only focus on objects and their neighbour relationships as illustrated in Figure~\ref{fig:heir_gcn}(a), the function of this layer is to preserve relationship edge embeddings for each object node. After the graph propagation during local GCN, the edge features will be integrated into the updated node embeddings. We discuss the details of local GCN step-by-step as follows:
	
	Given the scene graph $ \mathcal{G} = (\mathcal{V},  \mathcal{E})$, we firstly represent the directed graph as a bipartite graph, so each directed edge can be represented by two \textit{in-} and \textit{out-} undirected edges:
	\begin{equation}
		\mathcal{E}_{undirect} = [\mathcal{E}_{in}; \mathcal{E}_{out}],
	\end{equation}
	For example, after transforming to undirected graph, the directed edge 
	$  \langle man \xrightarrow[\text{}]{\text{wear}} shirt \rangle  $
	becomes two undirected edges $ [( man - wear_{in}); ( wear_{out} - shirt)]  $. Therefore, the \textit{directed} scene graph can be represented in the GCN.
	
	Furthermore, we still want to preserve edge with features in the proposed local GCN.
	Therefore, we further define the relationship edges as relationship nodes to enable GCN to utilise edge features. However, if we directly mix object nodes and relationship nodes together in a full graph, the model will get confused.
	As a result, we deploy local GCN to only focus on neighbour relationship nodes locally, and then the object-object connections will be further embedded in global GCN.

Formally, the input local sub-graphs contains object nodes and their neighbour relationship nodes (both \textit{in} and \textit{out}), the adjacency matrix $A^{(1)}$ represents the connected relationship nodes w.r.t. the object nodes. The local GCN output the embedding of all the nodes:
\begin{equation}
	\mathcal{V}^{(1)} = \sigma(W_{gcn}^{(1)}\mathcal{V}_{local}A^{(1)}),\label{eq:localgcn}
\end{equation}
	where
	$r$ is the number of object nodes, $s$ is the number of relationship types,
	$\mathcal{V}_{local} \in \mathbb{R}^{d \times (r+2s) } $ contains all the $r$ object nodes and $2s$ \textit{in}-\textit{out}-relationship nodes, 
	$d$ is the dimension of node features.
	$W_{gcn}^{(1)} \in \mathbb{R}^{D_g \times d }$  is the parameter to be learned, where $D_g$ is the dimension of GCN embedding space, here we set $D_g = d $ for consistency.
And the adjacency matrix $A^{(1)} \in \mathbb{R}^{(r+2s) \times (r+2s) } $ preserves the graph structure, the link between un-connected objects is set to zero. $\sigma$ is the non-linear activation function (e.g. ReLU).

	Furthermore, during the graph propagation, for each node, the embeddings of neighbour nodes will be added to form the new node embeddings as shown in Eq.\ref{eq:localgcn}.
	
	Therefore, after the propagation, the $r$ \textbf{object} node will contain the information of its \textbf{relationship} neighbour nodes. After the local updating, we will only keep the updated $r$ \textbf{object} nodes for further global GCN calculation, the $2s$ edge nodes are discarded. We denote the refined object node embeddings as $\hat{\mathcal{V}}^{(1)} \in \mathbb{R}^{d \times r }$.

\textbf{(b) Global GCN}. Since the local GCN has already captured the neighbour relationships for each node, the relationship-aware nodes and their undirected connections between each other are embedded in the global GCN:
\begin{equation}
	\hat{\mathcal{V}} = \sigma(W_{gcn}^{(2)}\hat{\mathcal{V}}^{(1)}A^{(2)}),
\end{equation}
where $W_{gcn}^{(2)} \in \mathbb{R}^{D_g \times d }$  is parameter to be learned, and  $A^{(2)} \in \mathbb{R}^{r \times r } $ is the adjacency matrix for object nodes.

Finally, we obtained relationship-aware node embeddings $ \hat{\mathcal{V}} =  [\hat{v_1}, \hat{v_2}, \cdots, \hat{v_r}] $, where $ \hat{v_i} \in \mathbb{R}^{d} $ denotes the $d$-dim node embedding of each object node, $r$ is the number of objects nodes. If the object is absent in the image, the node embeddings will be set to zeros.

\subsubsection{Graph Attention for Visual Dialogue} \label{sec:method-graph-attn}
Traditionally, the visual attention~\cite{cap_xu2015show} is a mechanism to reason about the most salient visual regions where the language model need to focus on while word sequence generation. Inspired by the visual attention, we propose a graph attention module to infer which object nodes should be currently focusing on.

\begin{figure}[!htb]
	\includegraphics[width=0.5\textwidth, trim={0.5cm 0.8cm 0.5cm 0.8cm}]{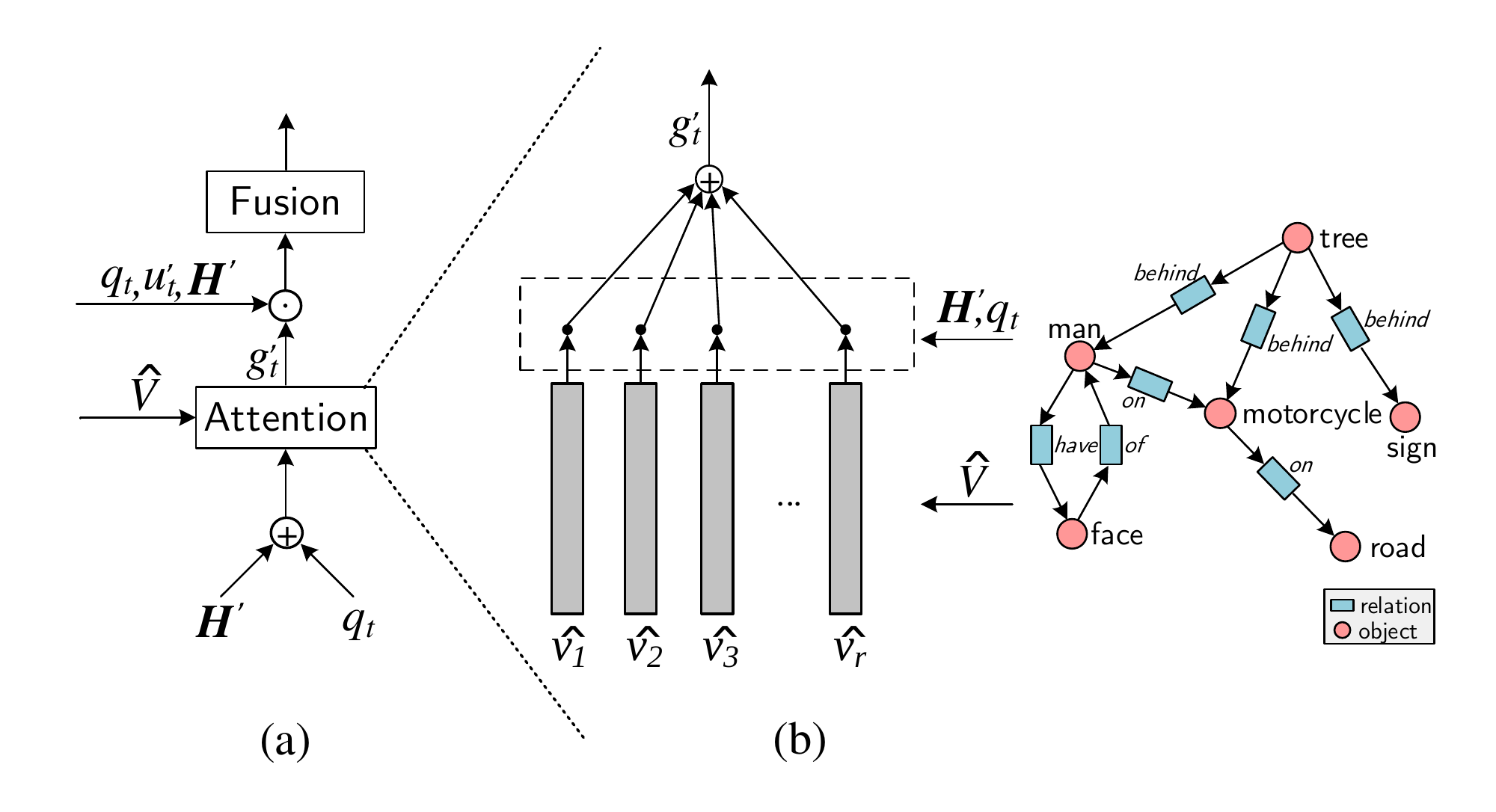}
	\centering
	\caption{An illustration of the graph attention module.}
	\label{fig:graph_attn_detail}
\end{figure}

In the graph attention model (Figure~\ref{fig:graph_attn_detail}),
given the learned graph embeddings $ \hat{\mathcal{V}} $, the graph attention attends to different relationship-aware object nodes given current states to provide context-aware scene graph representation.

Concretely, we use the relationship-aware graph embeddings $ \hat{\mathcal{V}} \in\mathbb{R}^{d\times r} $ from the proposed HierGCN in Section~\ref{sec:method-hierGCN}, the question feature $q_t \in \mathbb{R}^{d}$ at round $t$ generated by LSTM, 
and the attended dialogue history embeddings $\mathbf{H^{\prime}} \in \mathbb{R}^{d}$ following Section~\ref{sec:method-coattn}.

Given $ \hat{\mathcal{V}} \in\mathbb{R}^{d\times r} $, the context features  $q_t \in \mathbb{R}^{d}$ and $\mathbf{H^{\prime}} \in \mathbb{R}^{d}$, we feed them through a fully-connected (FC) neural network layer. The outputs from FC layer are passed through a softmax function to obtain the attention scores over $r$ object nodes:
\begin{equation}
	\begin{split}
		z_t^g &= w_a^T \tanh(W_{v}\hat{\mathcal{V}}+(W_{q} q_t) \mathbbm{1} ^T + (W_{h} \mathbf{H^{\prime}}) \mathbbm{1} ^T),\\
		\alpha_t^g &= \text{softmax}(z_t^g),
	\end{split}
\end{equation}
where $\mathbbm{1} \in \mathbb{R}^{r}$ is a vector filled with 1 to repeat current state to match the size of nodes. $W_{v}$, $W_{h} \in \mathbb{R}^{D_a\times d}$ denotes a latent $D_a$-dimensional embedding space for dimension reduction, $w_a^T \in \mathbb{R}^{D_a}$ are model parameters to learn. $\alpha_t^g \in \mathbb{R}^{r}$ is the attention score over the object $r$ node features. The context vector (attended graph embeddings) can be further calculated as a weighted sum of all the nodes:
\begin{equation}
	g_t^{\prime} = \sum_{i=1}^{r} \alpha_{ti}^g \hat{v_i},
\end{equation}
where the context vector $g_t^{\prime}$ is forwarded and combined with other features into encoder feature $E_t$.

\subsection{Co-Attention Network Encoder} \label{sec:method-coattn}
The co-attention strategy~\cite{vqa_HieCoAtt_LuYBP16, vd_nips_LuKYPB17, vd_cvpr_Wu0S0H18} is commonly adopted for reasoning the correlation between multiple feature inputs.

In this section, we introduce a visual relationship aware co-attention structure (Figure~\ref{fig:overview_system}) to solve the co-reference resolution between $ I $, $ \mathcal{G} $, $ \mathbf{H} $ and $ q_t $, and encode all the features into a combined feature representation for final response generation.

First, we recapitulate the unattended (raw) feature embeddings of the image, scene graph, question and history.
In terms of image representation, we use the top-$k$ region features $ \mathbf{U} \in \mathbb{R}^{D_u \times k}$ extracted from $I$ introduced in Section~\ref{sec:method-preliminary}. 
The relationship-aware graph node embeddings $\hat{\mathcal{V}} \in \mathbb{R}^{d \times r}$ with $r$ object nodes is further embedded following the proposed HierGCN (Section~\ref{sec:method-hierGCN}).
For textual content, the question at round $t$ is encoded with a standard language model (e.g. LSTM) to get the feature vector $q_t \in  \mathbb{R}^{d} $. Similarly, the previous history is encoded with another history language model as $\mathbf{H} \in \mathbb{R}^{d \times t}$.
Given question $q_t$, we firstly reason which history dialogues are relevant to current question. The history attention predicts the attention scores, and the $t$ dialogue embeddings are weighted summed based on the attention scores obtaining the attended history $\mathbf{H^{\prime}} \in \mathbb{R}^{d}$.
For example, given a question ``Does \textit{it} look old or new?'', it is difficult to precisely understand the meaning of pronoun \textit{it} directly. But by reasoning history dialogue, we can effortlessly locate the object from the previous dialogue ``What colour is the \textit{bike}?''.

Moreover, the image representation $ \mathbf{U} $ contains top-$k$ object regions, but not all the visual regions are related to the current question and attended history. The visual attention swiftly shift focus on salient object regions to decide which part of the regions are mostly important. Intuitively, if the questioner asks ``What is he wearing?'' in Figure~\ref{fig:overview_showcase}, the visual attention scores should reflect more weights on the \textit{shirt} region. We use linear layer to reduce the region feature from $D_u$ to $d$ dimension. The final weighted sum of region features is denoted as $ \mathbf{U^{\prime}} \in \mathbb{R}^{d} $.

Furthermore, the proposed graph attention (Section~\ref{sec:method-graph-attn}) outputs the attended graph embedding $g_t^{\prime}$ to preserve the visual relationships.

Finally, $\mathbf{H^{\prime}}$,$ \mathbf{U^{\prime}}$, $g_t^{\prime}$ and $q_t$  are concatenated and fused through a fully-connected fusion layer into encoder output $E_t$:
\begin{equation}
	E_t = \tanh(W_{f}[\mathbf{H^{\prime}}; \mathbf{U^{\prime}}; g_t^{\prime}; q_t]),
\end{equation}
where $W_{f} \in \mathbb{R}^{d \times 4d}$. The encoder output is then forwarded to discriminative and generative decoders respectively.


\section{Experiments} \label{sec:experiment}
\subsection{Experimental Settings}
\subsubsection{Dataset}
We conduct the experiments on the benchmark dataset - Visual Dialog dataset v0.9~\cite{vd_cvpr17_das}, which is created for the visual dialogue generation task. 
The training set contains 82,783 images-dialogue pairs, the validation set has 40,504 pairs from MSCOCO~\cite{mscoco}. Each dialogue contains 10 rounds of question-answer pairs. Besides, the pronouns (e.g. ``he'', ``she'', ``it'') frequently appear in the dataset, i.e. 98\% dialogues, 38\% of questions, and 19\% of answers contain at least one pronoun, unavoidable causing the co-reference resolution problems.
\subsubsection{Implementation Details}
Given an image, we extract region features using Faster-RCNN with ResNet-101~\cite{He_2016_CVPR} following the process in~\cite{vqa_updown}, the top-36 regions with highest confidence are selected for 2048-dimensional region feature representations. The scene graph is generated by pre-trained Factorizable Net~\cite{sg_fnet_2018}, and the top-50 detected predicates are chosen for scene graph construction. The language models used in question and history encoding are single-layer LSTM units with the hidden size of 512. The dimension of word embeddings is empirically set to 300, with random initialisation. For optimisation, we use stochastic optimisation method Adam~\cite{adam_KingmaB14} with learning rate of 4e-4 and coefficients ranging from 0.8 to 0.999. The batch size is fixed at 128. All the experiments are tested on a server with a 40-Core Intel(R) Xeon(R) E5-2660 CPU and 2 Nvidia GeForce GTX 1080 Ti GPUs.

\subsubsection{Evaluation Metrics}
We report the performance of dialogue generation using the common evaluation protocols from~\cite{vd_cvpr17_das, vd_nips_LuKYPB17, vd_cvpr_Wu0S0H18}. In testing, the visual dialogue model is given an image, a question, the multi-round dialogue history and a list of $N$ candidate answers ($N$=100), and the model will return a sort of the candidate answers. The returned ranked list of answers is evaluated on retrieval metrics \textit{w.r.t} human responses: Mean Reciprocal Rank (MRR), Recall$@$k (k=1, 5, 10), and Mean Rank (MR).
Both the discriminative and generative dialogue response decoders are compatible with the retrieval setting. Specifically, the discriminative decoder directly scores the confidence of candidate answers, whilst the generative model gives log-likelihood scores of each candidate answer.

\subsection{Compared Methods}\label{sec:experiment-compare-baseline}
The experiments on both discriminative (\textbf{-D}) and generative (\textbf{-G}) decoders are conducted.
In this section, we briefly introduce the compared visual dialogue methods. 

\textbf{SAN-QI-D~\cite{vqa_SAN_YangHGDS16}}: This Stacked Attention Networks (SAN) baseline is a VQA model, given only the question and image (without dialogue history) for answer prediction. 

\textbf{HieCoAtt-QI-D~\cite{vqa_HieCoAtt_LuYBP16}}: The Hierarchical Question-Image Co-Attention model utilises visual and hierarchical representation of the question in a joint framework for VQA.

\textbf{LF~\cite{vd_cvpr17_das}}: The Late Fusion (LF) encodes the image, question and dialogue history in three separate streams. Next, the model simply combines the image features, question embeddings and history embeddings into a joint representation for answer prediction.

\textbf{HREA~\cite{vd_cvpr17_das}}: In addition to LF model, the Hierarchical Recurrent Encoder with Attention model (HREA) adopts a hierarchical recurrent model to encode dialogue history in hierarchy equipped with a history attention.

\textbf{MN~\cite{vd_cvpr17_das}}: Memory Network is applied in visual dialogue task to enhance historical questions and answers memorising.

\textbf{HCIAE~\cite{vd_nips_LuKYPB17}}: The History-Conditioned Image Attentive Encoder (HCIAE) introduces an attentive framework to localise image regions, relevant history given the target question for co-reference resolution.
	For discriminative setting, we compare with HCIAE with MLE loss (-MLE), and with N-pair discriminative loss and self-attentive answer encoder (-NP-ATT) variants. While for generative setting, we mainly focus on HCIAE trained on maximum likelihood estimation (-MLE) loss, since the knowledge transfer (-DIS) is not in the scope of our method.

\textbf{CoAtt~\cite{vd_cvpr_Wu0S0H18}}: The Sequential Co-attention (CoAtt) encoder attends to each input feature by other features sequentially to capture the co-relation between all the input features. We compare our method with the sequential co-attention encoder with the MLE objective, without any adversarial learning strategies. The result of CoAtt discriminative (CoAtt-D) model with MLE objective is not available in the original paper, so the reported scores are based on our implementation. We also compare with their best model equipped with adversarial loss, intermediate reward and teacher forcing strategy (-GAN-TF).

	\textbf{GNN-SPO~\cite{vd_Zheng_2019_CVPR}}: Reasoning Visual Dialogs with Structural and Parital Observations (GNN-SPO). GNN-SPO utilises a graph structure to represent the dialogue, in which the nodes are dialogue entities, while the edges are semantic dependencies between the nodes. This method considers the dialogue history as partial observation of the graph, and the target is to infer the values of unobserved answer nodes and the graph structure.

\subsection{Quantitative Analysis}\label{sec:quantitative}
	
\begin{table}[!b]
	\centering
	\small
	\caption{Discriminative - Performance on Visual Dialog validation dataset~\cite{vd_cvpr17_das}}
	\label{tab:comparison_d}
	\begin{tabular}{l*{6}{c}}
		\toprule
		Model 		& MRR$\uparrow$ & R$@$1$\uparrow$ & R$@$5$\uparrow$ & R$@$10$\uparrow$ & MR$\downarrow$\\
		\midrule
		
		SAN-QI-D~\cite{vqa_SAN_YangHGDS16} & 0.5764 & 43.44 & 74.26 & 83.72 & 5.88\\
		HieCoAtt-QI-D~\cite{vqa_HieCoAtt_LuYBP16} & 0.5788 & 43.51 & 74.49 & 83.96 & 5.84\\
		LF-D~\cite{vd_cvpr17_das} &0.5807 & 43.82 & 74.68 & 84.07 & 5.78\\
		HREA-D~\cite{vd_cvpr17_das}  &0.5868 & 44.82 & 74.81 & 84.36 & 5.66\\
		MN-D~\cite{vd_cvpr17_das}  &0.5965 & 45.55 & 76.22 & 85.37 & 5.46\\
		
		HCIAE-D-MLE~\cite{vd_nips_LuKYPB17} & 0.6140 & 47.73 & 77.50 & 86.35 & 5.15\\ 
		HCIAE-D-NP-ATT~\cite{vd_nips_LuKYPB17} & 0.6222 & 48.48 & 78.75 & 87.59 & 4.81 \\
		CoAtt-D-MLE~\cite{vd_cvpr_Wu0S0H18} & 0.6135 & 47.49 & 77.92 & 86.75 & 5.04\\ 
		CoAtt-D-GAN~\cite{vd_cvpr_Wu0S0H18} & 0.6398 & 50.29 & 80.71 & 88.81 & 4.47\\ 
		GNN-SPO~\cite{vd_Zheng_2019_CVPR} & 0.6285 & 48.95 & 79.65 & 88.36 & 4.57\\
		\midrule
		
		
		ORD-D \textsubscript{w/{ }SG}& 0.6340 & 49.93 & 79.70 & 88.20 & 4.66\\ 
		ORD-D \textsubscript{w/{ }SG+Rel}& 0.6383 & 50.46 & 80.09 & 88.54 & 4.56\\ 
		\textbf{ORD-D \textsubscript{w/{ }SG+Rel+Attn}} & \textbf{0.6447} & \textbf{51.22} & \textbf{80.67} & \textbf{89.01} & \textbf{4.44}\\ 
		
		\bottomrule
	\end{tabular}

\end{table}

\begin{table}[!b]
	\centering
	\small
	\caption{Generative - Performance on Visual Dialog validation dataset~\cite{vd_cvpr17_das}}
	\label{tab:comparison_g}
	\begin{tabular}{l*{6}{c}}
		\toprule
		Model 		& MRR$\uparrow$ & R$@$1$\uparrow$ & R$@$5$\uparrow$ & R$@$10$\uparrow$ & MR$\downarrow$\\
		\midrule
		
		LF-G~\cite{vd_cvpr17_das}  & 0.5199 & 41.83 & 61.78 & 67.59 & 17.07\\
		HREA-G~\cite{vd_cvpr17_das}  & 0.5242 & 42.28 & 62.33 & 68.17 & 16.79\\
		MN-G~\cite{vd_cvpr17_das}  & 0.5259  & 42.29  & 62.85 & 68.88 & 17.06\\
		
		HCIAE-G-MLE~\cite{vd_nips_LuKYPB17} & 0.5382 & 44.07 & 63.42 & 69.03 & 16.06\\
		HCIAE-G-DIS~\cite{vd_nips_LuKYPB17} & 0.5467 & 44.35 & 65.28 & 71.55 & 14.23\\
		CoAtt-G-MLE~\cite{vd_cvpr_Wu0S0H18} & 0.5411 & 44.32 & 63.82 & 69.75 & 16.47\\ 
		CoAtt-G-GAN-TF~\cite{vd_cvpr_Wu0S0H18} & 0.5578 & 46.10 & 65.69 & 71.74 & 14.43\\
		\midrule
		
		
		ORD-G \textsubscript{w/{ }SG} & 0.5438 & 45.05 & 63.52 & 69.01 & 16.17 \\ 
		ORD-G  \textsubscript{w/{ }SG+Rel} &0.5480 & 45.47 & 64.04 & 69.59 & 15.83\\ 
		\textbf{ORD-G  \textsubscript{w/{ }SG+Rel+Attn}} & \textbf{0.5502} & \textbf{45.63} & \textbf{64.38} & \textbf{70.23} & \textbf{15.56}\\ 
		\bottomrule
	\end{tabular}
\end{table}

The main results on discriminative setting on the VisDial dataset are demonstrated in Table~\ref{tab:comparison_d}. In general, the proposed ORD full model outperforms most of state-of-the-art on all the metrics demonstrating the effectiveness of leveraging scene graph. Comparing to LF-D, ORD significantly improves MRR and Recall$@$1 by relatively 11\% and 17\%, respectively. This observation indicates that visual reasoning and co-attention are critical to dialogue generation.

	Moreover, the scene graph improves the answer prediction by a significant margin comparing to co-reference resolution methods (e.g. MN, HCIAE, CoAtt).
	Specifically, ORD outperforms HCIAE-D-NP-ATT by relatively 4\% on MRR from 0.6222 to 0.6447 
	confirming the visual relationships can enhance the model to recognise visual cues for dialogue reasoning. Besides, similar trend can be found in recall,
	the proposed method also makes a 6\% relative improvement on Recall$@$1. 
	We can also observe that the fine-grained object regions and relationships outperforms the GNN-based method built on the dialogue entities graph (e.g. GNN-SPO). In this regard, the graph structure provides strong reasoning performance where the underlying relationships are important.

	Furthermore, we reports the performance of ORD and the compared methods under the generative setting in Table~\ref{tab:comparison_g}. In all the metrics, ORD performs the best among other compared methods, especially outperforming the previous state-of-the-art CoAtt-G-MLE by relative 2\% and 3\% in terms of MRR and Recall$@$1.
	This shows that the object relationships are well preserved in the graph-structured multi-modal sources encoding framework to benefit the dialogue generator.
	
	The performance of CoAtt-G-GAN-TF is slightly higher than ours because the GAN and teacher-forcing strategy. However, since those training strategies are not the main focus of this work, therefore they are not in the scope of our method. Theoretically, these training strategies are applicable into our framework to further improve the final result.

\subsection{Ablative evaluation} \label{sec:experiment-ablation}
	In this section, we present several variants of our proposed method to study the contribution of each component in our ORD framework. Two baseline models LF and HCIAE-MLE are included for reference. The details of the variants are described as follows:
	
	\textbf{ORD\textsubscript{w/o CoAttn}}: This model keeps visual features, scene graph features, question embeddings, and history features without Co-Attention mechanism between all of them. It is similar to LF~\cite{vd_cvpr17_das}, but with region-level features and scene graph.
	
	\textbf{ORD\textsubscript{w/o Vis}}: This model keeps only scene graph, question and histories without visual features.
	
	\textbf{ORD\textsubscript{w/o SG}}: The model contains only visual features, questions and histories without scene graph.
	
	\textbf{ORD\textsubscript{w/ SG}}: This model is the base model to leverage scene graph in high-level. It keeps only undirected link between object nodes, without fine-grained relationship edge features (e.g. \textit{man-shirt}, \textit{bus-tree}). The nodes are embedded via single layer GCN (global GCN). The average of node embeddings forms the scene graph feature.
	
	\textbf{ORD\textsubscript{w/ SG+Rel}}: This model keeps directed link between objects nodes, with relationship edges features (e.g. \textit{man-wear-shirt}, \textit{shirt-on-man}). The nodes and edges are embedded by the proposed Hierarchical GCN (Global+Local GCN). The average of relationship-aware node embeddings forms the scene graph feature.
	
	\textbf{ORD\textsubscript{w/ SG+Rel+Attn}}: Our full model. This model keeps directed link between objects nodes, with edges features (e.g. \textit{man-wear-shirt}, \textit{shirt-on-man}). The nodes and edges are embedded by the proposed Hierarchical GCN. The attended relationship-aware node embeddings forms the scene graph feature.
\begin{table}[!t]
	\centering
	\small
	\caption{Ablative Study - Discriminative Setting}
	\label{tab:ablative}
	\begin{tabular}{l*{6}{c}}
		\toprule
		Model 		& MRR$\uparrow$ & R$@$1$\uparrow$ & R$@$5$\uparrow$ & R$@$10$\uparrow$ & MR$\downarrow$\\
		\midrule
		LF~\cite{vd_cvpr17_das} &0.5807 & 43.82 & 74.68 & 84.07 & 5.78\\
		HCIAE-MLE~\cite{vd_nips_LuKYPB17} & 0.6140 & 47.73 & 77.50 & 86.35 & 5.15\\ 
		\midrule
		
		\textbf{ORD \textsubscript{w/o CoAttn}} & 0.6129 & 47.46 & 77.79 & 86.66 & 5.04\\
		\textbf{ORD \textsubscript{w/o Vis}} & 0.5990 & 45.75 & 76.47 & 85.46 & 5.40\\
		\textbf{ORD \textsubscript{w/o SG}} & 0.6266 & 49.10 & 79.08 & 87.79 & 4.77\\
		\textbf{ORD \textsubscript{w/{ }SG}} & 0.6340 & 49.93 & 79.70 & 88.20 & 4.66\\ 
		\textbf{ORD \textsubscript{w/{ }SG+Rel}} & 0.6383 & 50.46 & 80.09 & 88.54 & 4.56\\ 
		\textbf{ORD \textsubscript{w/{ }SG+Rel+Attn}} & \textbf{0.6447} & \textbf{51.22} & \textbf{80.67} & \textbf{89.01} & \textbf{4.44}\\ 
		
		\bottomrule
	\end{tabular}
\end{table}

	For discriminative setting, our base model ORD\textsubscript{w/ SG} outperforms the maximum likelihood trained (-MLE) models such as LF, HCIAE shown in Table \ref{tab:ablative}. 
	The result shows that the fine-grained object regions and connections between them can significantly improve the visual reasoning capability.
	\begin{figure*}[!t]
		\includegraphics[width=0.97\textwidth, trim={0.7cm 0.8cm 0.7cm 0.6cm}, clip=true]{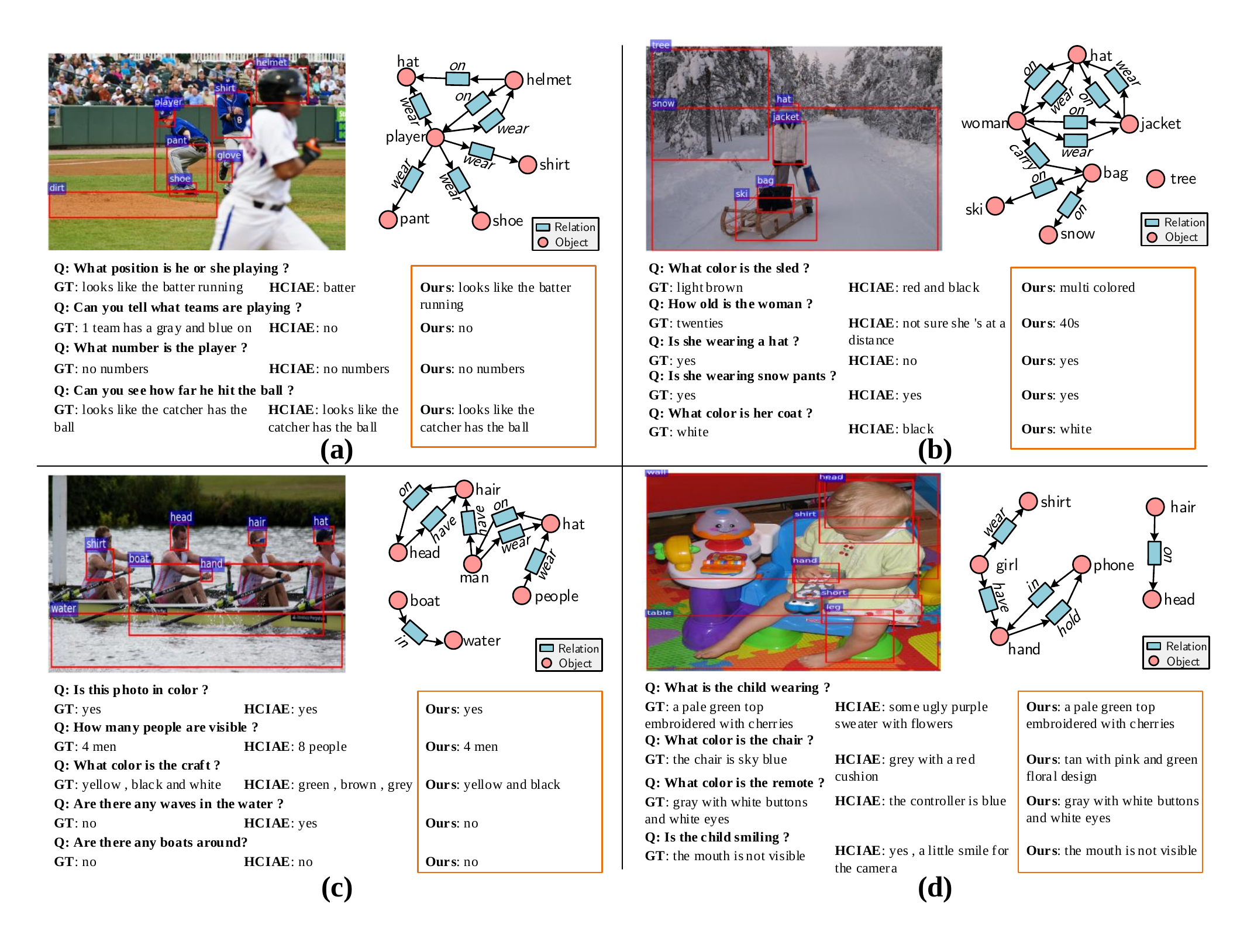}
		\centering
		\caption{Example visual dialogue results. Each example is associated with the ground-truth (GT), state-of-the-art method HCIAE~\cite{vd_nips_LuKYPB17} and the proposed ORD (ours).}
		\label{fig:casestudy}
	\end{figure*}
	Whilst this variant shows the effectiveness of the proposed Global GCN, we also want to evaluate the contribution of Local GCN. However, it is not reasonable to only use Local GCN without Global GCN. 
	The reason is that without linking object nodes via Global GCN layer, the nodes in the graph are not connecting to each other, so the nodes cannot pass message in the graph along edges. As a result, we add relationship edge features in addition to ORD\textsubscript{w/ SG}, so the ORD\textsubscript{w/  SG+Rel} can illustrate the additional contribution of local GCN. As we can see in Table~\ref{tab:ablative}, the fine-grained relationships (edge features) considered in ORD \textsubscript{w/ SG+Rel} improves the performance slightly, showing the Local GCN is effective to exploit the edge features.
	Furthermore, our full model ORD \textsubscript{w/ SG+Rel+Attn} further boosts the proposed scene graph-based reasoning framework and achieves 0.6447 on MRR, which demonstrates that the graph neural networks is able to dynamically focus on different semantic object nodes while answer reasoning.
	In addition, we remove co-attention, visual features and graph embeddings respectively to evaluate the importance of each the building blocks in the full framework.
	First, when we remove the co-attention module, the structure become to be very similar to LF model. However, in ORD\textsubscript{w/o CoAttn}, the additional region-level visual features and semantic scene graphs improves the performance. Nevertheless, the co-attention module is still an essential part of the model, otherwise the co-relation between different features cannot be well discovered.
	Second, in ORD\textsubscript{w/o Vis}, the removal of visual features degrades the performance dramatically, since the visual dialogue task is discussing the topics related to the given image. However, it is interesting that the MRR is still slightly better than LF model even no visual content is presented. 
	We conclude that the well-defined co-attention and scene graph are the major reasons in this scenario, because the co-attention is capturing underlying interactions between history conversations, while the scene graph is preserving the semantic object relationships from the visual content.
	Third, when the model ORD\textsubscript{w/o SG} completely removes scene graph, we can see the performance is better than HCIAE-MLE thanks to the fine-grained region features. 
	The structured visual relationships are indispensible for further improvement.
	
	For generative setting, the proposed full model outperforms most of the state-of-the-art models shown in Table~\ref{tab:comparison_g}. 
	However, the improvement of base model ORD-G\textsubscript{w/ SG} is not as significant as the discriminative model, which indicates that the sequential generative decoder needs further feature refinement strategy at sequence decoding stage. Surprisingly, the contribution of ORD-G\textsubscript{w/ SG+Rel} is slightly better than graph attention ORD-G\textsubscript{w/ SG+Rel+Attn} in generative model which is different from the trend in discriminative setting. We believe that how to optimise the graph attention in generative decoder should also be considered in the future work.


\subsection{Qualitative Analysis}

We present qualitative results from the proposed ORD and state-of-the-art HCIAE~\cite{vd_nips_LuKYPB17} in Figure~\ref{fig:casestudy}. The following properties can be observed from the results.

\textbf{Fine-grained visual cues.} The region visual features are preserved by integrating object detection in the visual dialogue framework, retaining detailed attributes of objects, which further enable the dialogue model to answer visually grounded questions (e.g. colour of objects). 
For example, in Figure~\ref{fig:casestudy}(b), the question is ``What colour is her coat?''. The dominant colour in this snow scene is white, and the coat is also white. It is a challenging question to CNN-based HCIAE model, since CNN features lose the object level details in visual representation. But by utilising the fine-grained region features, the colour of coat is easily observed by ORD model.

\textbf{Geometric object relationship.} By preserving the visual relationships in the scene graph, the ORD model can precisely reason the interactions between objects, which is omitted in previous work. Especially in the complex scene like Figure~\ref{fig:casestudy}(d) with complicated arranged colourful objects. Comparing to ground truth, the ORD inevitably have some failure cases such as recognising the colour of the chair. But when comparing to the state-of-the-art method, the proposed object relationship discovery model still remains strong performance. Take the first question as an example, the child is wearing a pale green top which is recognised by ORD seamlessly. Similarly, in Figure~\ref{fig:casestudy}(d), for the question ``Is she wearing a hat?'', since the $\langle \mathit{woman-wear-hat} \rangle$ is preserved in scene graph, the ORD is able to effortlessly predict the correct answer.

\textbf{Robust to background noise. }Because the ORD model recognises and preserves the objects in the visual encoding stream, the background noise can be separated from foreground salient objects. By taking the Figure~\ref{fig:casestudy}(c) as an example, the high reflection on the water surface is very noisy, making the HCIAE unable to recognise the number of people, the colour of craft, and the existence of waves. However, the proposed ORD is able to locate people, craft, and water regardless of background reflection and low lighting, therefore providing precise dialogue response in most of the cases.

\section{Conclusion and Future Work} \label{sec:conclusion}
In this paper, we propose a Object Relationship Discovery framework to generate fine-grained dialogue by discovering visual relationships from the image, where the detected locative and subtle object relationships assist the model to understand visual cues in an accurate way.
To preserve the discovered local and global object relationships, a hierarchical graph convolutional network is constructed followed by a graph attention. The graph attention module selectively focuses on relevant relationships, therefore eliminating background noise to ease visual evidence reasoning.
The experiments have demonstrated the superiority of our proposed method compared to the state-of-the-art. 
Furthermore, in other relevant tasks such as image captioning and VQA, the object relationships are also important cues for visual reasoning. Our future direction is to utilise visual relationships in other vision-language tasks.


%


\bibliographystyle{ACM-Reference-Format}
\bibliography{conf_bib}


\begin{thebibliography}{37}


\ifx \showCODEN    \undefined \def \showCODEN     #1{\unskip}     \fi
\ifx \showDOI      \undefined \def \showDOI       #1{#1}\fi
\ifx \showISBNx    \undefined \def \showISBNx     #1{\unskip}     \fi
\ifx \showISBNxiii \undefined \def \showISBNxiii  #1{\unskip}     \fi
\ifx \showISSN     \undefined \def \showISSN      #1{\unskip}     \fi
\ifx \showLCCN     \undefined \def \showLCCN      #1{\unskip}     \fi
\ifx \shownote     \undefined \def \shownote      #1{#1}          \fi
\ifx \showarticletitle \undefined \def \showarticletitle #1{#1}   \fi
\ifx \showURL      \undefined \def \showURL       {\relax}        \fi
\providecommand\bibfield[2]{#2}
\providecommand\bibinfo[2]{#2}
\providecommand\natexlab[1]{#1}
\providecommand\showeprint[2][]{arXiv:#2}

\bibitem[\protect\citeauthoryear{Anderson, He, Buehler, Teney, Johnson, Gould,
  and Zhang}{Anderson et~al\mbox{.}}{2018}]%
        {vqa_updown}
\bibfield{author}{\bibinfo{person}{Peter Anderson}, \bibinfo{person}{Xiaodong
  He}, \bibinfo{person}{Chris Buehler}, \bibinfo{person}{Damien Teney},
  \bibinfo{person}{Mark Johnson}, \bibinfo{person}{Stephen Gould}, {and}
  \bibinfo{person}{Lei Zhang}.} \bibinfo{year}{2018}\natexlab{}.
\newblock \showarticletitle{Bottom-Up and Top-Down Attention for Image
  Captioning and Visual Question Answering}. In \bibinfo{booktitle}{\emph{The
  IEEE Conference on Computer Vision and Pattern Recognition (CVPR)}}.
  \bibinfo{pages}{6077--6086}.
\newblock


\bibitem[\protect\citeauthoryear{Antol, Agrawal, Lu, Mitchell, Batra, Zitnick,
  and Parikh}{Antol et~al\mbox{.}}{2015}]%
        {vqa_iccv15_AntolALMBZP15}
\bibfield{author}{\bibinfo{person}{Stanislaw Antol}, \bibinfo{person}{Aishwarya
  Agrawal}, \bibinfo{person}{Jiasen Lu}, \bibinfo{person}{Margaret Mitchell},
  \bibinfo{person}{Dhruv Batra}, \bibinfo{person}{C.~Lawrence Zitnick}, {and}
  \bibinfo{person}{Devi Parikh}.} \bibinfo{year}{2015}\natexlab{}.
\newblock \showarticletitle{{VQA:} Visual Question Answering}. In
  \bibinfo{booktitle}{\emph{2015 {IEEE} International Conference on Computer
  Vision, {ICCV} 2015, Santiago, Chile, December 7-13, 2015}}.
  \bibinfo{pages}{2425--2433}.
\newblock


\bibitem[\protect\citeauthoryear{Bin, Yang, Zhou, Huang, and Shen}{Bin
  et~al\mbox{.}}{2017}]%
        {mm_imgcap_binyi}
\bibfield{author}{\bibinfo{person}{Yi Bin}, \bibinfo{person}{Yang Yang},
  \bibinfo{person}{Jie Zhou}, \bibinfo{person}{Zi Huang}, {and}
  \bibinfo{person}{Heng~Tao Shen}.} \bibinfo{year}{2017}\natexlab{}.
\newblock \showarticletitle{Adaptively Attending to Visual Attributes and
  Linguistic Knowledge for Captioning}. In \bibinfo{booktitle}{\emph{MM}}
  \emph{(\bibinfo{series}{MM '17})}. \bibinfo{publisher}{ACM},
  \bibinfo{address}{New York, NY, USA}, \bibinfo{pages}{1345--1353}.
\newblock
\showISBNx{978-1-4503-4906-2}
\urldef\tempurl%
\url{https://doi.org/10.1145/3123266.3123391}
\showDOI{\tempurl}


\bibitem[\protect\citeauthoryear{Choi, Chao, Pantofaru, and Savarese}{Choi
  et~al\mbox{.}}{2013}]%
        {sg_3d_ChoiCPS13}
\bibfield{author}{\bibinfo{person}{Wongun Choi}, \bibinfo{person}{Yu{-}Wei
  Chao}, \bibinfo{person}{Caroline Pantofaru}, {and} \bibinfo{person}{Silvio
  Savarese}.} \bibinfo{year}{2013}\natexlab{}.
\newblock \showarticletitle{Understanding Indoor Scenes Using 3D Geometric
  Phrases}. In \bibinfo{booktitle}{\emph{The IEEE Conference on Computer Vision
  and Pattern Recognition (CVPR)}}. \bibinfo{pages}{33--40}.
\newblock


\bibitem[\protect\citeauthoryear{Dai, Zhang, and Lin}{Dai
  et~al\mbox{.}}{2017}]%
        {sg_DaiZL17}
\bibfield{author}{\bibinfo{person}{Bo Dai}, \bibinfo{person}{Yuqi Zhang}, {and}
  \bibinfo{person}{Dahua Lin}.} \bibinfo{year}{2017}\natexlab{}.
\newblock \showarticletitle{Detecting Visual Relationships with Deep Relational
  Networks}. In \bibinfo{booktitle}{\emph{The IEEE Conference on Computer
  Vision and Pattern Recognition (CVPR)}}. \bibinfo{pages}{3298--3308}.
\newblock


\bibitem[\protect\citeauthoryear{Das, Kottur, Gupta, Singh, Yadav, Moura,
  Parikh, and Batra}{Das et~al\mbox{.}}{2017}]%
        {vd_cvpr17_das}
\bibfield{author}{\bibinfo{person}{Abhishek Das}, \bibinfo{person}{Satwik
  Kottur}, \bibinfo{person}{Khushi Gupta}, \bibinfo{person}{Avi Singh},
  \bibinfo{person}{Deshraj Yadav}, \bibinfo{person}{Jos{\'{e}} M.~F. Moura},
  \bibinfo{person}{Devi Parikh}, {and} \bibinfo{person}{Dhruv Batra}.}
  \bibinfo{year}{2017}\natexlab{}.
\newblock \showarticletitle{Visual Dialog}. In \bibinfo{booktitle}{\emph{The
  IEEE Conference on Computer Vision and Pattern Recognition (CVPR)}}.
  \bibinfo{pages}{1080--1089}.
\newblock


\bibitem[\protect\citeauthoryear{Gao, Zeng, Song, Li, Liu, Mei, and Shen}{Gao
  et~al\mbox{.}}{2019}]%
        {aaai_vqa_gao}
\bibfield{author}{\bibinfo{person}{Lianli Gao}, \bibinfo{person}{Pengpeng
  Zeng}, \bibinfo{person}{Jingkuan Song}, \bibinfo{person}{Yuan{-}Fang Li},
  \bibinfo{person}{Wu Liu}, \bibinfo{person}{Tao Mei}, {and}
  \bibinfo{person}{Heng~Tao Shen}.} \bibinfo{year}{2019}\natexlab{}.
\newblock \showarticletitle{Structured Two-Stream Attention Network for Video
  Question Answering}. In \bibinfo{booktitle}{\emph{AAAI}}.
  \bibinfo{pages}{6391--6398}.
\newblock


\bibitem[\protect\citeauthoryear{Gkioxari, Girshick, and Malik}{Gkioxari
  et~al\mbox{.}}{2015}]%
        {sg_action_GkioxariGM15}
\bibfield{author}{\bibinfo{person}{Georgia Gkioxari}, \bibinfo{person}{Ross~B.
  Girshick}, {and} \bibinfo{person}{Jitendra Malik}.}
  \bibinfo{year}{2015}\natexlab{}.
\newblock \showarticletitle{Contextual Action Recognition with R*CNN}. In
  \bibinfo{booktitle}{\emph{2015 {IEEE} International Conference on Computer
  Vision, {ICCV} 2015, Santiago, Chile, December 7-13, 2015}}.
  \bibinfo{pages}{1080--1088}.
\newblock


\bibitem[\protect\citeauthoryear{Gupta and Davis}{Gupta and Davis}{2008}]%
        {sg_GuptaD08}
\bibfield{author}{\bibinfo{person}{Abhinav Gupta} {and}
  \bibinfo{person}{Larry~S. Davis}.} \bibinfo{year}{2008}\natexlab{}.
\newblock \showarticletitle{Beyond Nouns: Exploiting Prepositions and
  Comparative Adjectives for Learning Visual Classifiers}. In
  \bibinfo{booktitle}{\emph{Computer Vision - {ECCV} 2008, 10th European
  Conference on Computer Vision, Marseille, France, October 12-18, 2008,
  Proceedings, Part {I}}}. \bibinfo{pages}{16--29}.
\newblock


\bibitem[\protect\citeauthoryear{Gupta, Kembhavi, and Davis}{Gupta
  et~al\mbox{.}}{2009}]%
        {sg_action_GuptaKD09}
\bibfield{author}{\bibinfo{person}{Abhinav Gupta}, \bibinfo{person}{Aniruddha
  Kembhavi}, {and} \bibinfo{person}{Larry~S. Davis}.}
  \bibinfo{year}{2009}\natexlab{}.
\newblock \showarticletitle{Observing Human-Object Interactions: Using Spatial
  and Functional Compatibility for Recognition}.
\newblock \bibinfo{journal}{\emph{{IEEE} Trans. Pattern Anal. Mach. Intell.}}
  \bibinfo{volume}{31}, \bibinfo{number}{10} (\bibinfo{year}{2009}),
  \bibinfo{pages}{1775--1789}.
\newblock


\bibitem[\protect\citeauthoryear{Han, Shen, Liu, Yang, and Shen}{Han
  et~al\mbox{.}}{2018}]%
        {mm_rel_HTShen}
\bibfield{author}{\bibinfo{person}{Chaojun Han}, \bibinfo{person}{Fumin Shen},
  \bibinfo{person}{Li Liu}, \bibinfo{person}{Yang Yang}, {and}
  \bibinfo{person}{Heng~Tao Shen}.} \bibinfo{year}{2018}\natexlab{}.
\newblock \showarticletitle{Visual Spatial Attention Network for Relationship
  Detection}. In \bibinfo{booktitle}{\emph{MM}}. \bibinfo{pages}{510--518}.
\newblock
\urldef\tempurl%
\url{https://doi.org/10.1145/3240508.3240611}
\showDOI{\tempurl}


\bibitem[\protect\citeauthoryear{He, Zhang, Ren, and Sun}{He
  et~al\mbox{.}}{2016}]%
        {He_2016_CVPR}
\bibfield{author}{\bibinfo{person}{Kaiming He}, \bibinfo{person}{Xiangyu
  Zhang}, \bibinfo{person}{Shaoqing Ren}, {and} \bibinfo{person}{Jian Sun}.}
  \bibinfo{year}{2016}\natexlab{}.
\newblock \showarticletitle{Deep Residual Learning for Image Recognition}. In
  \bibinfo{booktitle}{\emph{The IEEE Conference on Computer Vision and Pattern
  Recognition (CVPR)}}. \bibinfo{pages}{770--778}.
\newblock


\bibitem[\protect\citeauthoryear{Johnson, Gupta, and Fei{-}Fei}{Johnson
  et~al\mbox{.}}{2018}]%
        {sg_JohnsonGF18}
\bibfield{author}{\bibinfo{person}{Justin Johnson}, \bibinfo{person}{Agrim
  Gupta}, {and} \bibinfo{person}{Li Fei{-}Fei}.}
  \bibinfo{year}{2018}\natexlab{}.
\newblock \showarticletitle{Image Generation From Scene Graphs}. In
  \bibinfo{booktitle}{\emph{The IEEE Conference on Computer Vision and Pattern
  Recognition (CVPR)}}. \bibinfo{pages}{1219--1228}.
\newblock


\bibitem[\protect\citeauthoryear{Johnson, Krishna, Stark, Li, Shamma,
  Bernstein, and Li}{Johnson et~al\mbox{.}}{2015}]%
        {sg_JohnsonKSLSBL15}
\bibfield{author}{\bibinfo{person}{Justin Johnson}, \bibinfo{person}{Ranjay
  Krishna}, \bibinfo{person}{Michael Stark}, \bibinfo{person}{Li{-}Jia Li},
  \bibinfo{person}{David~A. Shamma}, \bibinfo{person}{Michael~S. Bernstein},
  {and} \bibinfo{person}{Fei{-}Fei Li}.} \bibinfo{year}{2015}\natexlab{}.
\newblock \showarticletitle{Image retrieval using scene graphs}. In
  \bibinfo{booktitle}{\emph{The IEEE Conference on Computer Vision and Pattern
  Recognition (CVPR)}}. \bibinfo{pages}{3668--3678}.
\newblock


\bibitem[\protect\citeauthoryear{Karpathy and Fei{-}Fei}{Karpathy and
  Fei{-}Fei}{2017}]%
        {cap_Karpathy}
\bibfield{author}{\bibinfo{person}{Andrej Karpathy} {and} \bibinfo{person}{Li
  Fei{-}Fei}.} \bibinfo{year}{2017}\natexlab{}.
\newblock \showarticletitle{Deep Visual-Semantic Alignments for Generating
  Image Descriptions}.
\newblock \bibinfo{journal}{\emph{{IEEE} Trans. Pattern Anal. Mach. Intell.}}
  \bibinfo{volume}{39}, \bibinfo{number}{4} (\bibinfo{year}{2017}),
  \bibinfo{pages}{664--676}.
\newblock


\bibitem[\protect\citeauthoryear{Kingma and Ba}{Kingma and Ba}{2015}]%
        {adam_KingmaB14}
\bibfield{author}{\bibinfo{person}{Diederik~P. Kingma} {and}
  \bibinfo{person}{Jimmy Ba}.} \bibinfo{year}{2015}\natexlab{}.
\newblock \showarticletitle{Adam: {A} Method for Stochastic Optimization}. In
  \bibinfo{booktitle}{\emph{3rd International Conference on Learning
  Representations, {ICLR} 2015, San Diego, CA, USA, May 7-9, 2015, Conference
  Track Proceedings}}.
\newblock


\bibitem[\protect\citeauthoryear{Kipf and Welling}{Kipf and Welling}{2017}]%
        {gcn_KipfW17}
\bibfield{author}{\bibinfo{person}{Thomas~N. Kipf} {and} \bibinfo{person}{Max
  Welling}.} \bibinfo{year}{2017}\natexlab{}.
\newblock \showarticletitle{Semi-Supervised Classification with Graph
  Convolutional Networks}. In \bibinfo{booktitle}{\emph{5th International
  Conference on Learning Representations, {ICLR} 2017, Toulon, France, April
  24-26, 2017, Conference Track Proceedings}}.
\newblock


\bibitem[\protect\citeauthoryear{Kottur, Moura, Parikh, Batra, and
  Rohrbach}{Kottur et~al\mbox{.}}{2018}]%
        {vd_kottur2018visual}
\bibfield{author}{\bibinfo{person}{Satwik Kottur}, \bibinfo{person}{Jos{\'e}~MF
  Moura}, \bibinfo{person}{Devi Parikh}, \bibinfo{person}{Dhruv Batra}, {and}
  \bibinfo{person}{Marcus Rohrbach}.} \bibinfo{year}{2018}\natexlab{}.
\newblock \showarticletitle{Visual coreference resolution in visual dialog
  using neural module networks}. In \bibinfo{booktitle}{\emph{Computer Vision -
  {ECCV} 2018 - 15th European Conference, Munich, Germany, September 8-14,
  2018, Proceedings}}. \bibinfo{pages}{153--169}.
\newblock


\bibitem[\protect\citeauthoryear{Krause, Johnson, Krishna, and
  Fei{-}Fei}{Krause et~al\mbox{.}}{2017}]%
        {cap_krause2016paragraphs}
\bibfield{author}{\bibinfo{person}{Jonathan Krause}, \bibinfo{person}{Justin
  Johnson}, \bibinfo{person}{Ranjay Krishna}, {and} \bibinfo{person}{Li
  Fei{-}Fei}.} \bibinfo{year}{2017}\natexlab{}.
\newblock \showarticletitle{A Hierarchical Approach for Generating Descriptive
  Image Paragraphs}. In \bibinfo{booktitle}{\emph{The IEEE Conference on
  Computer Vision and Pattern Recognition (CVPR)}}.
  \bibinfo{pages}{3337--3345}.
\newblock


\bibitem[\protect\citeauthoryear{Li, Ouyang, Bolei, Jianping, Chao, and
  Wang}{Li et~al\mbox{.}}{2018}]%
        {sg_fnet_2018}
\bibfield{author}{\bibinfo{person}{Yikang Li}, \bibinfo{person}{Wanli Ouyang},
  \bibinfo{person}{Zhou Bolei}, \bibinfo{person}{Shi Jianping},
  \bibinfo{person}{Zhang Chao}, {and} \bibinfo{person}{Xiaogang Wang}.}
  \bibinfo{year}{2018}\natexlab{}.
\newblock \showarticletitle{Factorizable Net: An Efficient Subgraph-based
  Framework for Scene Graph Generation}. In \bibinfo{booktitle}{\emph{Computer
  Vision - {ECCV} 2018 - 15th European Conference, Munich, Germany, September
  8-14, 2018, Proceedings, Part {I}}}. \bibinfo{pages}{346--363}.
\newblock


\bibitem[\protect\citeauthoryear{Li, Ouyang, Zhou, Wang, and Wang}{Li
  et~al\mbox{.}}{2017}]%
        {sg_2017_li_msdn}
\bibfield{author}{\bibinfo{person}{Yikang Li}, \bibinfo{person}{Wanli Ouyang},
  \bibinfo{person}{Bolei Zhou}, \bibinfo{person}{Kun Wang}, {and}
  \bibinfo{person}{Xiaogang Wang}.} \bibinfo{year}{2017}\natexlab{}.
\newblock \showarticletitle{Scene Graph Generation from Objects, Phrases and
  Region Captions}. In \bibinfo{booktitle}{\emph{{IEEE} International
  Conference on Computer Vision, {ICCV} 2017, Venice, Italy, October 22-29,
  2017}}. \bibinfo{pages}{1270--1279}.
\newblock


\bibitem[\protect\citeauthoryear{Lin, Maire, Belongie, Hays, Perona, Ramanan,
  Doll{\'{a}}r, and Zitnick}{Lin et~al\mbox{.}}{2014}]%
        {mscoco}
\bibfield{author}{\bibinfo{person}{Tsung{-}Yi Lin}, \bibinfo{person}{Michael
  Maire}, \bibinfo{person}{Serge~J. Belongie}, \bibinfo{person}{James Hays},
  \bibinfo{person}{Pietro Perona}, \bibinfo{person}{Deva Ramanan},
  \bibinfo{person}{Piotr Doll{\'{a}}r}, {and} \bibinfo{person}{C.~Lawrence
  Zitnick}.} \bibinfo{year}{2014}\natexlab{}.
\newblock \showarticletitle{Microsoft {COCO:} Common Objects in Context}. In
  \bibinfo{booktitle}{\emph{Computer Vision - {ECCV} 2014 - 13th European
  Conference, Zurich, Switzerland, September 6-12, 2014, Proceedings, Part
  {V}}}. \bibinfo{pages}{740--755}.
\newblock


\bibitem[\protect\citeauthoryear{Lu, Krishna, Bernstein, and Li}{Lu
  et~al\mbox{.}}{2016a}]%
        {sg_LuKBL16}
\bibfield{author}{\bibinfo{person}{Cewu Lu}, \bibinfo{person}{Ranjay Krishna},
  \bibinfo{person}{Michael~S. Bernstein}, {and} \bibinfo{person}{Fei{-}Fei
  Li}.} \bibinfo{year}{2016}\natexlab{a}.
\newblock \showarticletitle{Visual Relationship Detection with Language
  Priors}. In \bibinfo{booktitle}{\emph{Computer Vision - {ECCV} 2016 - 14th
  European Conference, Amsterdam, The Netherlands, October 11-14, 2016,
  Proceedings, Part {I}}}. \bibinfo{pages}{852--869}.
\newblock


\bibitem[\protect\citeauthoryear{Lu, Kannan, Yang, Parikh, and Batra}{Lu
  et~al\mbox{.}}{2017}]%
        {vd_nips_LuKYPB17}
\bibfield{author}{\bibinfo{person}{Jiasen Lu}, \bibinfo{person}{Anitha Kannan},
  \bibinfo{person}{Jianwei Yang}, \bibinfo{person}{Devi Parikh}, {and}
  \bibinfo{person}{Dhruv Batra}.} \bibinfo{year}{2017}\natexlab{}.
\newblock \showarticletitle{Best of Both Worlds: Transferring Knowledge from
  Discriminative Learning to a Generative Visual Dialog Model}. In
  \bibinfo{booktitle}{\emph{Advances in Neural Information Processing Systems
  30: Annual Conference on Neural Information Processing Systems 2017, 4-9
  December 2017, Long Beach, CA, {USA}}}. \bibinfo{pages}{313--323}.
\newblock


\bibitem[\protect\citeauthoryear{Lu, Yang, Batra, and Parikh}{Lu
  et~al\mbox{.}}{2016b}]%
        {vqa_HieCoAtt_LuYBP16}
\bibfield{author}{\bibinfo{person}{Jiasen Lu}, \bibinfo{person}{Jianwei Yang},
  \bibinfo{person}{Dhruv Batra}, {and} \bibinfo{person}{Devi Parikh}.}
  \bibinfo{year}{2016}\natexlab{b}.
\newblock \showarticletitle{Hierarchical Question-Image Co-Attention for Visual
  Question Answering}. In \bibinfo{booktitle}{\emph{Advances in Neural
  Information Processing Systems 29: Annual Conference on Neural Information
  Processing Systems 2016, December 5-10, 2016, Barcelona, Spain}}.
  \bibinfo{pages}{289--297}.
\newblock


\bibitem[\protect\citeauthoryear{Niu, Zhang, Zhang, Zhang, Lu, and Wen}{Niu
  et~al\mbox{.}}{2019}]%
        {vd_niu2019recursive}
\bibfield{author}{\bibinfo{person}{Yulei Niu}, \bibinfo{person}{Hanwang Zhang},
  \bibinfo{person}{Manli Zhang}, \bibinfo{person}{Jianhong Zhang},
  \bibinfo{person}{Zhiwu Lu}, {and} \bibinfo{person}{Ji-Rong Wen}.}
  \bibinfo{year}{2019}\natexlab{}.
\newblock \showarticletitle{Recursive visual attention in visual dialog}. In
  \bibinfo{booktitle}{\emph{The IEEE Conference on Computer Vision and Pattern
  Recognition (CVPR)}}. \bibinfo{pages}{6679--6688}.
\newblock


\bibitem[\protect\citeauthoryear{Ren, He, Girshick, and Sun}{Ren
  et~al\mbox{.}}{2017}]%
        {faster_rcnn}
\bibfield{author}{\bibinfo{person}{Shaoqing Ren}, \bibinfo{person}{Kaiming He},
  \bibinfo{person}{Ross~B. Girshick}, {and} \bibinfo{person}{Jian Sun}.}
  \bibinfo{year}{2017}\natexlab{}.
\newblock \showarticletitle{Faster {R-CNN:} Towards Real-Time Object Detection
  with Region Proposal Networks}.
\newblock \bibinfo{journal}{\emph{{IEEE} Trans. Pattern Anal. Mach. Intell.}}
  \bibinfo{volume}{39}, \bibinfo{number}{6} (\bibinfo{year}{2017}),
  \bibinfo{pages}{1137--1149}.
\newblock


\bibitem[\protect\citeauthoryear{Teney, Liu, and van~den Hengel}{Teney
  et~al\mbox{.}}{2017}]%
        {vqa_graph_TeneyLH17}
\bibfield{author}{\bibinfo{person}{Damien Teney}, \bibinfo{person}{Lingqiao
  Liu}, {and} \bibinfo{person}{Anton van~den Hengel}.}
  \bibinfo{year}{2017}\natexlab{}.
\newblock \showarticletitle{Graph-Structured Representations for Visual
  Question Answering}. In \bibinfo{booktitle}{\emph{The IEEE Conference on
  Computer Vision and Pattern Recognition (CVPR)}}.
  \bibinfo{pages}{3233--3241}.
\newblock


\bibitem[\protect\citeauthoryear{Vinyals, Toshev, Bengio, and Erhan}{Vinyals
  et~al\mbox{.}}{2015}]%
        {cap_VinyalsTBE15}
\bibfield{author}{\bibinfo{person}{Oriol Vinyals}, \bibinfo{person}{Alexander
  Toshev}, \bibinfo{person}{Samy Bengio}, {and} \bibinfo{person}{Dumitru
  Erhan}.} \bibinfo{year}{2015}\natexlab{}.
\newblock \showarticletitle{Show and tell: {A} neural image caption generator}.
  In \bibinfo{booktitle}{\emph{The IEEE Conference on Computer Vision and
  Pattern Recognition (CVPR)}}. \bibinfo{pages}{3156--3164}.
\newblock


\bibitem[\protect\citeauthoryear{Wang, Luo, Li, Huang, and Yin}{Wang
  et~al\mbox{.}}{2018}]%
        {cap_depth_WangLLHY18}
\bibfield{author}{\bibinfo{person}{Ziwei Wang}, \bibinfo{person}{Yadan Luo},
  \bibinfo{person}{Yang Li}, \bibinfo{person}{Zi Huang}, {and}
  \bibinfo{person}{Hongzhi Yin}.} \bibinfo{year}{2018}\natexlab{}.
\newblock \showarticletitle{Look Deeper See Richer: Depth-aware Image Paragraph
  Captioning}. In \bibinfo{booktitle}{\emph{2018 {ACM} Multimedia Conference on
  Multimedia Conference, {MM} 2018, Seoul, Republic of Korea, October 22-26,
  2018}}. \bibinfo{pages}{672--680}.
\newblock


\bibitem[\protect\citeauthoryear{Wu, Wang, Shen, Reid, and van~den Hengel}{Wu
  et~al\mbox{.}}{2018}]%
        {vd_cvpr_Wu0S0H18}
\bibfield{author}{\bibinfo{person}{Qi Wu}, \bibinfo{person}{Peng Wang},
  \bibinfo{person}{Chunhua Shen}, \bibinfo{person}{Ian~D. Reid}, {and}
  \bibinfo{person}{Anton van~den Hengel}.} \bibinfo{year}{2018}\natexlab{}.
\newblock \showarticletitle{Are You Talking to Me? Reasoned Visual Dialog
  Generation Through Adversarial Learning}. In \bibinfo{booktitle}{\emph{The
  IEEE Conference on Computer Vision and Pattern Recognition (CVPR)}}.
  \bibinfo{pages}{6106--6115}.
\newblock


\bibitem[\protect\citeauthoryear{Xu, Zhu, Choy, and Fei{-}Fei}{Xu
  et~al\mbox{.}}{2017}]%
        {sg_XuZCF17}
\bibfield{author}{\bibinfo{person}{Danfei Xu}, \bibinfo{person}{Yuke Zhu},
  \bibinfo{person}{Christopher~B. Choy}, {and} \bibinfo{person}{Li Fei{-}Fei}.}
  \bibinfo{year}{2017}\natexlab{}.
\newblock \showarticletitle{Scene Graph Generation by Iterative Message
  Passing}. In \bibinfo{booktitle}{\emph{The IEEE Conference on Computer Vision
  and Pattern Recognition (CVPR)}}. \bibinfo{pages}{3097--3106}.
\newblock


\bibitem[\protect\citeauthoryear{Xu, Ba, Kiros, Cho, Courville, Salakhutdinov,
  Zemel, and Bengio}{Xu et~al\mbox{.}}{2015}]%
        {cap_xu2015show}
\bibfield{author}{\bibinfo{person}{Kelvin Xu}, \bibinfo{person}{Jimmy Ba},
  \bibinfo{person}{Ryan Kiros}, \bibinfo{person}{Kyunghyun Cho},
  \bibinfo{person}{Aaron~C. Courville}, \bibinfo{person}{Ruslan Salakhutdinov},
  \bibinfo{person}{Richard~S. Zemel}, {and} \bibinfo{person}{Yoshua Bengio}.}
  \bibinfo{year}{2015}\natexlab{}.
\newblock \showarticletitle{Show, Attend and Tell: Neural Image Caption
  Generation with Visual Attention}. In \bibinfo{booktitle}{\emph{Proceedings
  of the 32nd International Conference on Machine Learning, Lille, France}}.
  \bibinfo{pages}{2048--2057}.
\newblock


\bibitem[\protect\citeauthoryear{Yang, He, Gao, Deng, and Smola}{Yang
  et~al\mbox{.}}{2016}]%
        {vqa_SAN_YangHGDS16}
\bibfield{author}{\bibinfo{person}{Zichao Yang}, \bibinfo{person}{Xiaodong He},
  \bibinfo{person}{Jianfeng Gao}, \bibinfo{person}{Li Deng}, {and}
  \bibinfo{person}{Alexander~J. Smola}.} \bibinfo{year}{2016}\natexlab{}.
\newblock \showarticletitle{Stacked Attention Networks for Image Question
  Answering}. In \bibinfo{booktitle}{\emph{The IEEE Conference on Computer
  Vision and Pattern Recognition (CVPR)}}. \bibinfo{pages}{21--29}.
\newblock


\bibitem[\protect\citeauthoryear{Yao and Li}{Yao and Li}{2010}]%
        {sg_action_YaoF10}
\bibfield{author}{\bibinfo{person}{Bangpeng Yao} {and}
  \bibinfo{person}{Fei{-}Fei Li}.} \bibinfo{year}{2010}\natexlab{}.
\newblock \showarticletitle{Grouplet: {A} structured image representation for
  recognizing human and object interactions}. In \bibinfo{booktitle}{\emph{The
  IEEE Conference on Computer Vision and Pattern Recognition (CVPR)}}.
  \bibinfo{pages}{9--16}.
\newblock


\bibitem[\protect\citeauthoryear{Yao, Pan, Li, and Mei}{Yao
  et~al\mbox{.}}{2018}]%
        {cap_sg_YaoPLM18}
\bibfield{author}{\bibinfo{person}{Ting Yao}, \bibinfo{person}{Yingwei Pan},
  \bibinfo{person}{Yehao Li}, {and} \bibinfo{person}{Tao Mei}.}
  \bibinfo{year}{2018}\natexlab{}.
\newblock \showarticletitle{Exploring Visual Relationship for Image
  Captioning}. In \bibinfo{booktitle}{\emph{Computer Vision - {ECCV} 2018 -
  15th European Conference, Munich, Germany, September 8-14, 2018, Proceedings,
  Part {XIV}}}. \bibinfo{pages}{711--727}.
\newblock


\bibitem[\protect\citeauthoryear{Zheng, Wang, Qi, and Zhu}{Zheng
  et~al\mbox{.}}{2019}]%
        {vd_Zheng_2019_CVPR}
\bibfield{author}{\bibinfo{person}{Zilong Zheng}, \bibinfo{person}{Wenguan
  Wang}, \bibinfo{person}{Siyuan Qi}, {and} \bibinfo{person}{Song-Chun Zhu}.}
  \bibinfo{year}{2019}\natexlab{}.
\newblock \showarticletitle{Reasoning Visual Dialogs With Structural and
  Partial Observations}. In \bibinfo{booktitle}{\emph{The IEEE Conference on
  Computer Vision and Pattern Recognition (CVPR)}}.
\newblock


\end{thebibliography}

\end{document}